\documentclass[11pt,technote,onecolumn]{IEEEtran}
\usepackage{cite,amsmath,graphicx}
\usepackage[caption=false,font=footnotesize]{subfig}
\usepackage{bm}
\usepackage{amssymb}
\usepackage{booktabs}
\usepackage{multirow}
\usepackage{algorithm,algorithmic}

\usepackage{epstopdf}
\usepackage{amsthm}
\usepackage{lineno}
\usepackage{mathrsfs}
\usepackage{amsmath}
\usepackage{txfonts}
\usepackage{graphicx}
\usepackage{caption}

\newtheorem{definition}{Definition}

\newtheorem{theorem}{Theorem}%[section]
\DeclareMathOperator*{\argmin}{argmin}

\ifCLASSINFOpdf
\else
\fi
\begin{document}
%
% paper title
% Titles are generally capitalized except for words such as a, an, and, as,
% at, but, by, for, in, nor, of, on, or, the, to and up, which are usually
% not capitalized unless they are the first or last word of the title.
% Linebreaks \\ can be used within to get better formatting as desired.
% Do not put math or special symbols in the title.
\title{Bayesian Low Rank Tensor Ring Model for Image Completion}

% author names and affiliations
% transmag papers use the long conference author name format.
\author{Zhen~Long,
Ce~Zhu,~\IEEEmembership{Fellow,~IEEE}, Jiani~Liu, Yipeng~Liu,~\IEEEmembership{Senior Member,~IEEE}}

\IEEEtitleabstractindextext{%
\begin{abstract}
Low rank tensor ring model is powerful for image completion which recovers missing entries in data acquisition and transformation. The recently proposed tensor ring (TR) based completion algorithms generally solve the low rank optimization problem by alternating least squares method with predefined ranks, which may easily lead to overfitting when the unknown ranks are set too large and only a few measurements are available. In this paper, we present a Bayesian low rank tensor ring model for image completion by automatically learning the low-rank structure of data. A multiplicative interaction model is developed for the low-rank tensor ring decomposition, where core factors are enforced to be sparse by assuming their entries obey Student-T distribution. Compared with most of the existing methods, the proposed one is free of parameter-tuning, and the TR ranks can be obtained by Bayesian inference. Numerical Experiments, including synthetic data, color images with different sizes and YaleFace dataset B with respect to one pose, show that the proposed approach outperforms state-of-the-art ones, especially in terms of recovery accuracy.
%Low rank tensor ring (TR) completion is a powerful tool to recover missing entries in multi-dimensional data acquisition and transformation. The TR based completion algorithm generally solves the low rank optimization problem by alternating least squares method with predefined ranks, which may easily lead to overfitting when the unknown ranks are set too large and only a few measurements are available. In this paper, we present a Bayesian low TR rank model for tensor completion by learning the low-rank structure of data. A multiplicative interaction model is developed for the low-rank decomposition, where core factors are enforced to be sparse by applying hierarchical prior, assuming hyper-parameters Gamma distributed. Compared with most of the existing methods, the proposed one is free of parameter-tuning, and the TR ranks can be obtained by Bayesian inference. Experiments on 3D (color image), 4D (YaleFace dataset B with respect to one pose) and 5D (light field image) data show that the proposed method outperforms the state-of-the-art ones in terms of recovery quality.
\end{abstract}

% Note that keywords are not normally used for peerreview papers.
}

% make the title area
\maketitle

% To allow for easy dual compilation without having to reenter the
% abstract/keywords data, the \IEEEtitleabstractindextext text will
% not be used in maketitle, but will appear (i.e., to be "transported")
% here as \IEEEdisplaynontitleabstractindextext when the compsoc
% or transmag modes are not selected <OR> if conference mode is selected
% - because all conference papers position the abstract like regular
% papers do.
\IEEEdisplaynontitleabstractindextext
% \IEEEdisplaynontitleabstractindextext has no effect when using
% compsoc or transmag under a non-conference mode.

% For peer review papers, you can put extra information on the cover
% page as needed:
% \ifCLASSOPTIONpeerreview
% \begin{center} \bfseries EDICS Category: 3-BBND \end{center}
% \fi
%
% For peerreview papers, this IEEEtran command inserts a page break and
% creates the second title. It will be ignored for other modes.
\IEEEpeerreviewmaketitle

% The very first letter is a 2 line initial drop letter followed
% by the rest of the first word in caps.
%
% form to use if the first word consists of a single letter:
% \IEEEPARstart{A}{demo} file is ....
%
% form to use if you need the single drop letter followed by
% normal text (unknown if ever used by the IEEE):
% \IEEEPARstart{A}{}demo file is ....
%
% Some journals put the first two words in caps:
% \IEEEPARstart{T}{his demo} file is ....
%
% Here we have the typical use of a "T" for an initial drop letter
% and "HIS" in caps to complete the first word.

\section{Introduction}
\label{Introduction}
\IEEEPARstart{T}{ensors}, which are multi-dimensional generalizations of matrices, provide a natural representation for multidimensional data. Exploring the internal structure of tensors could help us obtain more latent information for high-dimensional data processing. For example, a color video is a forth-order tensor, which allows its temporal and spatial correlation could be simultaneously investigated. Recently, tensor-based methods have attracted interests in image processing problems\cite{cichocki2015tensor, imaizumi2017tensor,wang2018classification, vasilache2018tensor,gupta2018shampoo,zhong2015discriminant,li2017parsimonious,LONG2019301,7080836}. Image completion is one of them, which recovers the missing entries during acquisition and transformation.
%\IEEEPARstart{T}{ensors}, which are multi-dimensional generalizations of matrices, provide a natural representation for multidimensional data. Exploring the internal structure of tensor can help us obtain more latent information for high-dimensional data processing. For example, a color video is a forth-order tensor, leading its temporal and spatial correlation can be simultaneously investigated. Recently, tensor based methods have attracted more interests in many fields, such as image processing \cite{cichocki2015tensor, imaizumi2017tensor}, machine learning \cite{wang2018classification, vasilache2018tensor, } and Quantum Physics \cite{grisafi2018symmetry, qi2018higher}. In most cases, the entries of acquired multidimensional data are not complete and the problem of recovering missing data is regarded as tensor completion.

In recent works, the tensor-based methods of image completion are mostly addressed by assuming the data is low rank and mainly divided into two groups. One is based on the rank minimization model and can be presented as:
\begin{equation}
\min_{\mathcal{X}}\quad \operatorname{rank}(\mathcal{X})\quad\text{s. t.}\quad \mathcal{X}_{\mathbb{O}}=\mathcal{T}_{\mathbb{O}},
\end{equation}
where $\mathcal{X}\in \mathbb{R}^{I_{1}\times I_{2}\times \cdots \times I_{N}}$ is the estimated tensor, $\mathcal{T}$ is an $N$-th order observed tensor with the same size as $\mathcal{X}$, and $\mathbb{O}$ is an index set with the entries observed. Traditional decompositions based on rank minimization model such as Tucker decomposition\cite{tucker1966some,kolda2009tensor} and tensor-Singular Value Decomposition (t-SVD)~\cite{braman2010third,kilmer2013third} have been well studied. Rank minimization models are developed by minimizing the summation of nuclear norm regularization terms which correspond to the unfolding matrices of $\mathcal{X}$.
For example, Tucker ranks minimization is firstly proposed in~\cite{liu2013tensor}, followed by tubal rank minimization~\cite{zhang2014novel,zhang2017exact}. Besides, some improvements for these rank minimization methods are proposed in~\cite{xie2016volume,mu2014square,liu2016generalized,hu2017twist,xue2018low,zhou2018tensor}. Recently, an advanced tensor network named tensor train (TT) has been proposed in~\cite{oseledets2011tensor} and showed its advantage to capture the latent information for image completion~\cite{bengua2017efficient}. Moreover, as one of the generation formats of TT decomposition, tensor tree ranks minimization for image completion is proposed in~\cite{liu2019image}.

The other important class is factorization based methods which alternatively update the factor with the predefined rank. Mathematically, the low-rank tensor completion problem can be formulated as:
\begin{equation}
\min_{\mathcal{X}}\quad \frac{1}{2}\|\mathcal{P}_{\mathbb{O}}(\mathcal{X}-\mathcal{T})\|_{\text{F}}^{2}\quad\text{s. t.} \quad \operatorname{rank}(\mathcal{X})=R.
\end{equation}
where $\mathcal{P}_{\mathbb{O}}$ is the random sampling operator. In \cite{leurgans1993decomposition}, the model with CANDECOMP/PARAFAC (CP) rank known in advance is proposed and solved by the alternating least squares (ALS) algorithm. Besides, considering the factorization based model and ALS approach, some works such as HOOI~\cite{kolda2009tensor} with Tucker ranks, Tubal-ALS~\cite{liu2016low} with tubal rank, TT-ALS~\cite{grasedyck2015variants} with TT ranks and TR-ALS~\cite{wang2017efficient} with tensor ring (TR) ranks are investigated to fill these fields. Aparting from the ALS framework, Riemannian optimization scheme with nonlinear conjugate gradient approach has been explored to tackle the factorization based tensor completion model, resulting in CP-WOPT~\cite{acar2011scalable}, RMTC~\cite{kasai2016low}, RTTC~\cite{steinlechner2016riemannian}, TR-WOPT~\cite{yuan2018higher} and HTTC~\cite{da2015optimization}.
%Unfortunately, overfitting easily occurs in this group when rank is set to be large and a few observations are available.

In these methods, the advanced tensor networks such as TT and TR perform better than the CP/Tucker based methods in image completion because they can capture more correlations than the traditional tensor decompositions. Compared with TT, TR has more balanced and smaller ranks due to its  ring structure, which may be beneficial to explore more latent structure. However, it is intractable to directly minimize tensor ring rank since its corresponding fold matrix is hard to find due to its circular dimensional permutation invariance. The existing TR based methods are based on the ranks pre-defined in advance, which may easily lead to overfitting when rank is set to be large and a few observations are available.
 %Futhermore, Riemannian structure is one of the branches to solve the low-rank model with predefined rank, resulting in CP weighted optimization (CPWOPT)~\cite{acar2011scalable}, RMTC~\cite{kasai2016low}, TT-WOPT~\cite{yuan2018high} and TR-WOPT~\cite{yuan2018higher}.

 %the CP decomposition with BI model is firstly introduced in~\cite{zhao2015bayesian} without outliers. To naturally deal with outliers, Bayesian robust tensor completion for solving tensor completion is presented~\cite{zhao2016bayesian}. Besides, in~\cite{rai2014scalable}, a scalable Bayesian low CP rank decomposition is proposed to solve both continuous and binary data. Followed by~\cite{zhao2015} which updates the Tucker factor in BI framework.  T
%Motivated by these, we present a Bayesian inference model for solving low rank TR decomposition on image completion. Bayesian inference (BI), which models the low-rank problem in Bayesian formulation, shows a success in low-rank matrix completion~\cite{babacan2012sparse, yang2018fast} by automatically adjusting the tradeoff between rank and fitting error. In addition, some tensor based works~\cite{zhao2016bayesian, rai2014scalable, he2018knowledge} reveal the superiority of BI framework on low-rank tensor completion.
%To the best of our knowledge, this is the first work to investigate low rank TR model for image completion on Bayesian inference framework.
Motivated by these, we present a Bayesian inference (BI) model for inferring TR ranks to solve image completion problem. BI, which models the low-rank problem, shows a success in low-rank matrix factorization~\cite{babacan2012sparse,zhao2015l,shi2017rank, yang2018fast} by automatically adjusting the tradeoff between rank and fitting error. In addition, some tensor based works~\cite{zhao2016bayesian, rai2014scalable, he2018knowledge} reveal the superiority of BI framework on low-rank tensor completion. To the best of our knowledge, this is the first work to investigate low rank TR model for image completion on BI framework. Our objective is to infer the missing entries from observations by low rank tensor ring completion, while TR ranks can be determined automatically.

\begin{figure}[htbp]
\vskip -0.2in
\centering
\subfloat[A natural image]{
\begin{minipage}[t]{0.25\linewidth}
\centering
\includegraphics[width=0.9in]{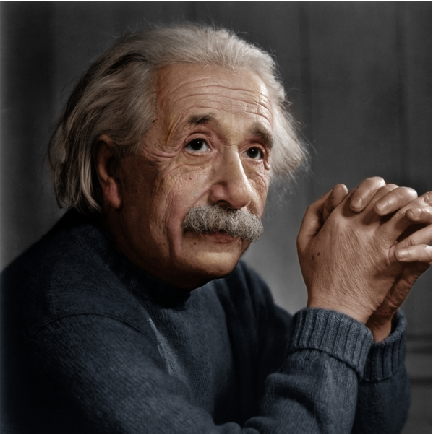}
%\caption{fig1}
\end{minipage}%
}%
\subfloat[derivation at horizontal]{
\begin{minipage}[t]{0.35\linewidth}
\centering
\includegraphics[width=1.3in]{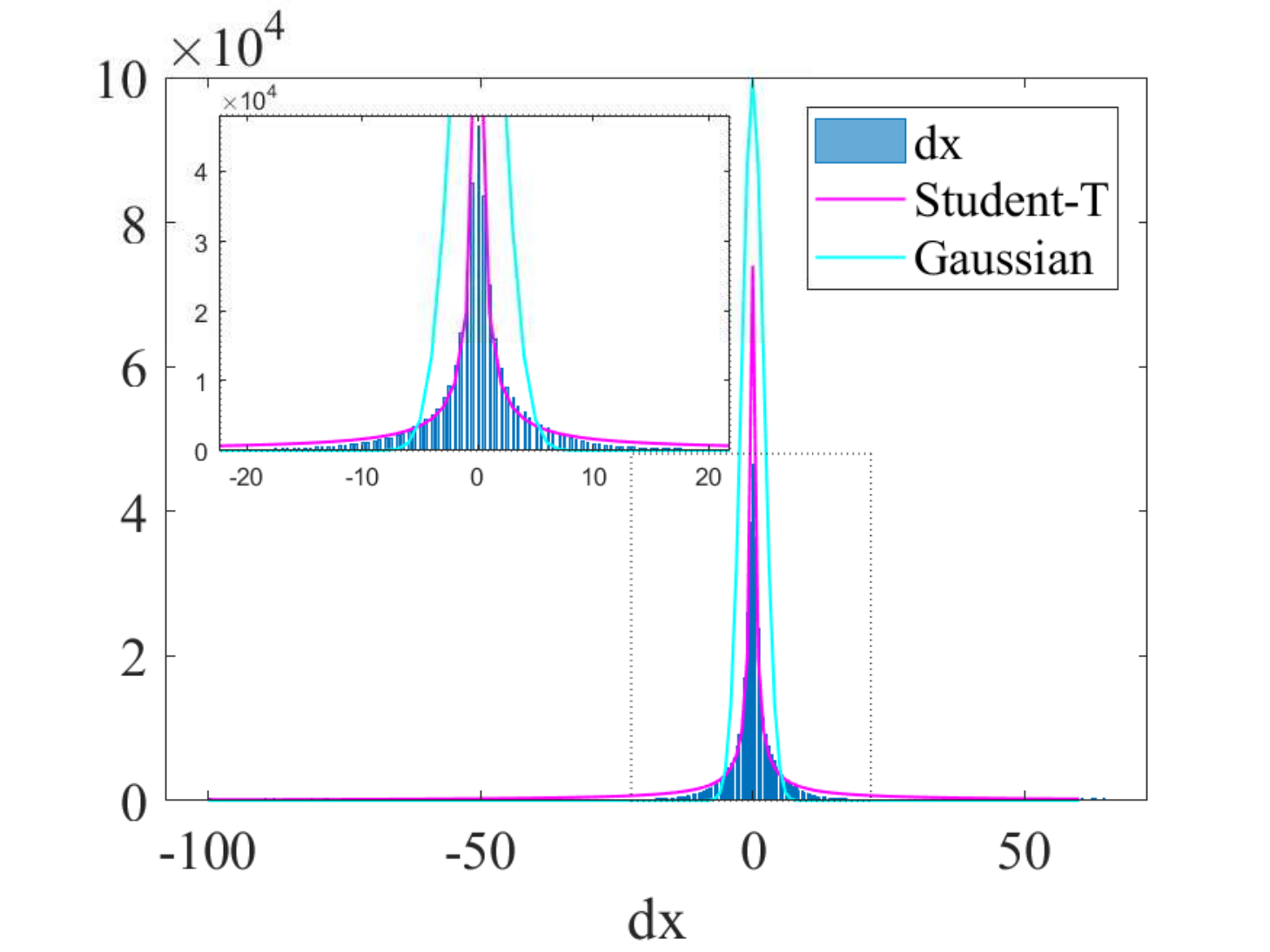}
%\caption{fig2}
\end{minipage}%
}%
\subfloat[derivation at vertical]{
\begin{minipage}[t]{0.35\linewidth}
\centering
\includegraphics[width=1.3in]{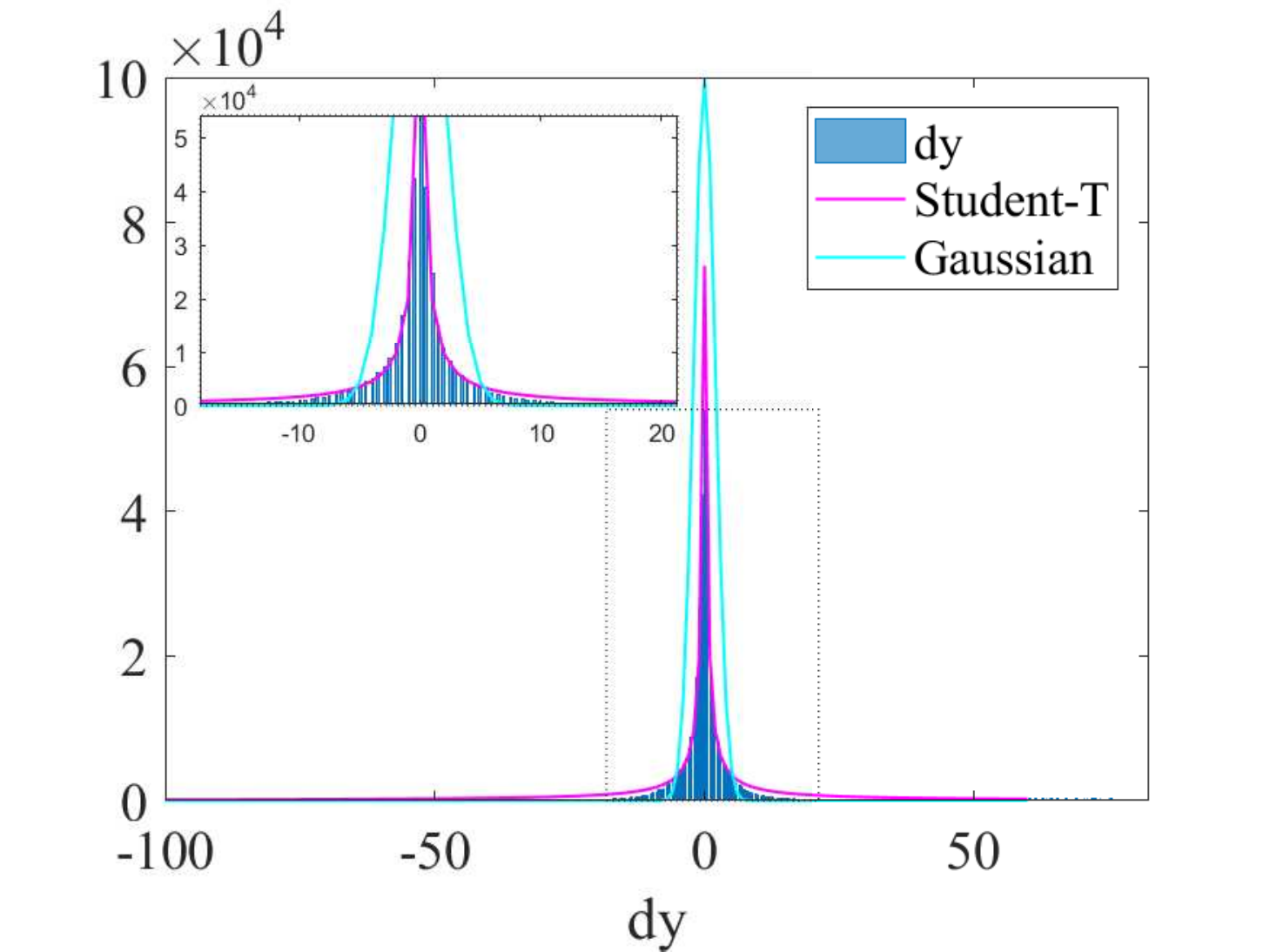}
%\caption{fig2}
\end{minipage}
}%
\centering
\caption{The property of heavy-tailed for a natural image}
\label{natural}
\end{figure}

The natural image has the characteristic of heavy-tailed. The phenomenon could be explained intuitively in Fig. \ref{natural}.
This property inspires us to assume the core factor, which is the potential part in TR formats, following Student-T distribution. To model this problem, we propose TR decomposition framework with a sparsity-inducing hierarchical prior over core factor. Specifically, each slice of core factor is assumed to independently follow a Gaussian distribution with zero mean and a variance matrix. The variance matrix is treated as a random hyperparameter with Gamma distribution. Then, variational BI algorithm has been utilized to estimate parameters in this model. Experiments on synthetic data and real-world images demonstrate that the recovery performance of the proposed approach outperforms existing state-of-the-art works, which may imply our algorithm can explore more correlations from images. In addition, the experiments also indicate that TR ranks inferred by our algorithm can be used in TR-ALS, which avoids tuning parameters.

 The rest of this paper is organized as follows. Sec. \ref{sec:2} introduces some notations and preliminaries for TR decomposition. In sec. \ref{sec:3}, the details of Bayesian TR decomposition are introduced, including the model description, solution and discussion. Sec. \ref{sec:4} provides some experiments on multi-way data completion. The conclusion is concluded in sec. \ref{sec:5}.

%In this paper, we proposed a new model with CGTN for tensor completion, which can explore more inter relationship between each mode of tensor. Besides, with the dimensional increase, the CGTN will be over-fit to the data we need to recover. Considering this, we adopt the similar ideas of Drop-out from deep learning~\cite{bibid}. The proposed optimization model can be solved by ALS which fixes one variable with others fixed.  Experimental results based on  synthetic data and color images synthetic data and color images that our method outperforms state-of-the-art algorithms in terms of peak signal-to-noise ratio (PSNR), structural similarity index (SSIM), and CPU time.
%Compared with the existing literature based on factorization, the main contributions of our work can be summarized as follows:

%(1) We propose a new CGTN model for tensor completion. It can capture more correlations than other tensor decomposition such as Tucker, CP, TT and TR.

%(2) With the dimension increase, the missing data with high G-rank may be overfit. We apply the drop-out function to reduce the drawback. Besides, TT and TR model can be special case of CGTN when drop-out the relationship.

\section{Notations and Preliminaries}
\label{sec:2}
\subsection{Notations}
Firstly we give the notations to be used. A scalar, a vector, a matrix, and a tensor are written as $x$, $ \mathbf{x} $,  $  \mathbf{X} $, and $\mathcal{X}$, respectively. The product of two scalars denotes $z=x*y$, $ \mathcal{X} \in \mathbb{R}^{ I_{1} \times\cdots\times I_{N}}$ denotes an $ N $-order tensor where $I_{n}$ is the dimensional size. The trace operation is denoted as $\operatorname{Trace}(\mathbf{X})=\sum_{i=1}^{I}x_{ii}$ where $\mathbf{X}\in \mathbb{R}^{I\times I}$ is a square matrix.  $\operatorname{Vec}({\mathcal{X}})$ denotes the vectorization of tensor $\mathcal{X}$  and $\operatorname{Ten}(\mathcal{X})\in \mathbb{R}^{ I_{1} \times\cdots\times I_{N}}$ denotes a tensor transforming from a vector $\mathbf{x}\in \mathbb{R}^{ I_{1} \cdots I_{N}}$ or a matrix $\mathbf{X}\in \mathbb{R}^{ I_{n} \times \prod_{j=1,j\neq n}^{N}I_{j}}$.
%$\langle{\mathcal X, \mathcal Y}\rangle =\langle \operatorname{Vec}(\mathcal{X}),\operatorname{Vec}(\mathcal{Y})\rangle  $ denotes the tensor inner product. The Frobenius norm of $\mathcal{ A} $ is denoted as $\lVert{\mathcal A}\rVert_{\text F}$ =$\langle{\mathcal A,\mathcal A}\rangle^\frac{1}{2} $.
The Kronecker product of two tensors can be denoted as $\mathcal{Z}=\mathcal{X}\otimes\mathcal{Y}\in\mathbb{R}^{I_{1}J_{1}\times\cdots\times I_{N}J_{N}}$, where $\mathcal{X}\in\mathbb{R}^{I_{1}\times\cdots\times I_{N}}$, $\mathcal{Y}\in \mathbb{R}^{J_{1}\times\cdots\times J_{N}}$.
The Hadamard product of two tensors is defined as $(\mathcal{X} \odot \mathcal{Y})_{i_1, \cdots , i_N}=\mathcal{X}_{i_1, \cdots , i_N}\mathcal{Y}_{i_1, \cdots , i_N}$, where $ {\mathcal{X}}_{i_1, \cdots , i_N} $ and $ {\mathcal{Y}}_{i_1, \cdots , i_N} $ are the entries of $ {\mathcal{X}} $ and $ {\mathcal{Y}} $.
%The mode-$n$ unfolding of $\mathcal{X}\in \mathbb{R}^{I_{1} \times\cdots\times I_{N}}$ is defined as $\mathcal{X}_{<n>}\in \mathbb{R}^{I_{n+1}\cdots I_{N}I_{1}\cdots I_{n-1}\times I_{n}}$.
%The mide-$n$ tensor-matrix product can be presented as $\mathcal{C}=\mathcal{A}\times_{n}\mathbf{B}$, where $\mathcal{A}\in \mathbb{R}^{I_{1} \times\cdots\times I_{N}}$, $\mathbf{B}\in \mathbb{R}^{J\times I_{n}}$, $\mathcal{C}\in \mathbb{R}^{I_{1} \times\cdots\times I_{n-1}\times J\times I_{n+1} \times I_{N}}$.

\subsection{Tensor Ring Model}
\begin{figure}[ht]
\vskip -0.2in
\begin{center}
\centerline{\includegraphics[width=\columnwidth]{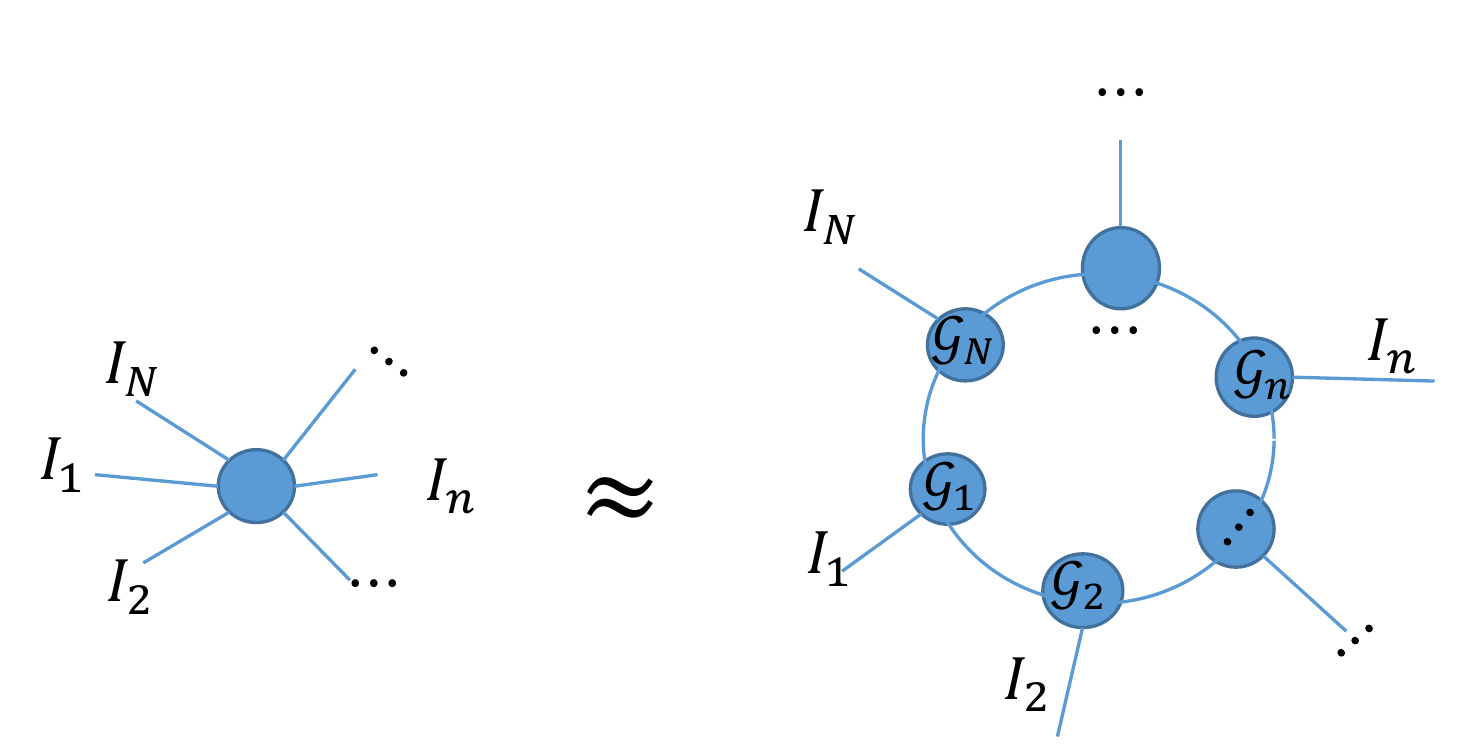}}
  \caption{TR decomposition}\label{Fig:TR}
\label{icml-historical}
\end{center}
\vskip -0.2in
\end{figure}

\begin{definition}
(\textbf{TR decomposition})~\cite{zhao2016tensor}
For an $N$-order tensor $\mathcal{X} \in \mathbb{R}^{I_{1} \times\cdots\times I_{N}}$, the TR decomposition is defined as
	\begin{eqnarray}\label{3}
	&&\mathcal{X}(i_1,i_2,\cdots, i_N)=\nonumber\\
    &&\operatorname{Trace}(\mathcal{G}_{1}(:,i_1,:)\mathcal{G}_{2}(:,i_2,:)\cdots \mathcal{G}_{N}(:,i_N,:)),
	\end{eqnarray}	
	where $\mathcal{G}_{n} \in \mathbb{R}^{R_{n-1}\times I_n\times R_{n}} $, $n=1, \cdots, N$ are the core factors, and the TR ranks are defined as $\{R_{n}, 0\leq n\leq N\}$ with $R_{0}=R_{N}$. For simplicity, we denote TR decomposition by $\mathcal{X}=f(\mathcal{G}_{1},\cdots,\mathcal{G}_{N})$. The graphical illustration of TR decomposition is shown in Fig. \ref{Fig:TR}.
\end{definition}

\begin{definition}
(\textbf{Tensor permutation})
For an $N$-order tensor $\mathcal{X} \in \mathbb{R}^{I_{1} \times\cdots\times I_{N}}$,  the tensor permutation is defined as $\mathcal{X}^{P_n} \in \mathbb{R}^{I_n\times \cdots\times   I_N\times I_1\cdots\times I_{n-1}}$:
	\begin{equation}
	\mathcal{X}^{P_n}(i_n,\cdots,i_N,i_1,\cdots,i_{n-1})=\mathcal{X}(i_1,\cdots,i_N)\nonumber.
	\end{equation}
\end{definition}
%	Then simple and advanced tensor network can be represented as $\mathcal{A}=f(\mathcal{U})=f(\mathcal{U}_{1}\mathcal{U}_{2}\cdots \mathcal{U}_{N})$, where function $f$ is a trace operation on $\mathcal{U}(:,i,:)$, $i=1, \cdots, I_1 I_2 \cdots I_N$, and followed by a reshaping operation from vector of the size $1\times(I_1 I_2 \cdots I_N)\times 1$ to tensor of the size $I_1\times I_2\times \cdots\times I_N$.	
\begin{theorem}
(\textbf{Cyclic permutation property})~\cite{zhao2016tensor}
Based on the definitions of tensor permutation and TR decomposition, the tensor permutation of $\mathcal{X}$ is equivalent to its factors circular shifting, as follows:
\begin{equation*}
	\mathcal{X}^{P_n}=f(\mathcal{G}_{n}, \cdots\mathcal{G}_{N}, \mathcal{G}_{1},\cdots\mathcal{G}_{n-1}),
	\end{equation*}
	with entries
	\begin{eqnarray}
	&\mathcal{X}^{P_n}(i_n,\cdots, i_N,i_1,\cdots, i_{n-1})=\operatorname{Trace}(\mathcal{G}_{n}(:,i_n,:)\nonumber\\
	&\cdots \mathcal{G}_{N}(:,i_N,:)\mathcal{G}_{1}(:,i_1,:)\cdots\mathcal{G}_{n-1}(:,i_{n-1},:)).\nonumber
	\end{eqnarray}
\end{theorem}
\begin{definition}
(\textbf{Tensor connection product (TCP)})~\cite{cichocki2014era}
The tensor connection product for 3-order tensors $\mathcal{G}_{n} \in \mathbb{R}^{R_{n-1} \times I_{n}\times R_{n}}$ is defined as
\begin{equation}
	\mathcal{G}=\operatorname{TCP}(\mathcal{G}_{1},\mathcal{G}_{2},\cdots, \mathcal{G}_{N})
  \in \mathbb{R}^{R_{0}\times (I_1\cdots I_N) \times R_{N}}\nonumber,
	\end{equation}		
and $\mathcal{G}^{\neq n}$, which is the TCP of a set of core factors excepting $\mathcal{G}_{n}$, is defined as:
\begin{eqnarray}
	\mathcal{G}^{\neq n}&=&\operatorname{TCP}(\mathcal{G}_{n+1},\cdots\mathcal{G}_{N}, \mathcal{G}_{1},\cdots, \mathcal{G}_{n-1})\nonumber \\
     &\in& \mathbb{R}^{R_{n}\times (I_{n+1}\cdots I_N I_1\cdots I_{n-1}) \times R_{n-1}}\nonumber.
	\end{eqnarray}	
\end{definition}
%Based on $\operatorname{TCP}$ definition, we can obtain a tensor $\mathcal{X} =f(\mathcal{G})\in \mathbb{R}^{I_{1}\times \cdots \times I_{N}}$ with entries $\mathcal{X}(i_1,\cdots, i_N)=\operatorname{Trace}(\mathcal{G}(:,i_{1}\cdots i_{N},:))$.

\begin{theorem}
(\textbf{Expectation of Inner product})
Letting a random tensor $\mathcal{X}=f(\mathcal{G}_{1}\mathcal{G}_{2}\cdots \mathcal{G}_{N})$, we can calculate the expectation of inner product by:
\begin{eqnarray}
&&\mathbb{E}[\langle\operatorname {Vec}(\mathcal{X}),\operatorname {Vec}(\mathcal{X})\rangle]\\\nonumber
&&=\sum_{i_{1}=1,\cdots,i_{N}}\prod_{n=1}^{N}(\mathbf{E}_{R_{n}}\otimes \mathbf{K}_{R_{n}R_{n-1}}\otimes \mathbf{E}_{R_{n-1}}) (\mathbb{E}[(\mathbf{g}_{n}(i_{n})(\mathbf{g}_{n}(i_{n}))^{\operatorname{T}}])
\end{eqnarray}
with
\begin{eqnarray}
&&\mathbb{E}[\operatorname{Vec}(\mathbf{g}_{n}(i_{n})))\operatorname{Vec}(\mathbf{g}_{n}(i_{n})))^{\operatorname{T}}]\\\nonumber
&&=\mathbb{E}[\operatorname{Vec}(\mathbf{g}_{n}(i_{n})))]\mathbb{E}[\operatorname{Vec}(\mathbf{g}_{n}(i_{n})))^{\operatorname{T}}]
+\text{Var}(\operatorname{Vec}(\mathbf{g}_{n}(i_{n})))).
\end{eqnarray}

\begin{proof}
\begin{eqnarray}
&&\mathbb{E}[\langle\operatorname {Vec}(\mathcal{X}),\operatorname {Vec}(\mathcal{X})\rangle]\nonumber\\
&&=\mathbb{E}[\sum_{i_{1}=1,\cdots,i_{N}}\mathcal{X}(i_1,\cdots, i_N)\mathcal{X}(i_1,\cdots, i_N)]\nonumber\\
&&=\mathbb{E}[\sum_{i_{1}=1,\cdots,i_{N}}\text{Trace}(\prod_{n=1}^{N}\mathbf{G}_{n}(i_{n}))\text{Trace}(\prod_{n=1}^{N}\mathbf{G}_{n}(i_{n}))]\nonumber\\
&&=\mathbb{E}[\sum_{i_{1}=1,\cdots,i_{N}}\text{Trace}((\prod_{n=1}^{N}\mathbf{G}_{n}(i_{n})))\otimes(\prod_{n=1}^{N}\mathbf{G}_{n}(i_{n}))))]\nonumber\\
&&=\mathbb{E}[\sum_{i_{1}=1,\cdots,i_{N}}\text{Trace}(\prod_{n=1}^{N}(\mathbf{G}_{n}(i_{n})\otimes\mathbf{G}_{n}(i_{n}))]\nonumber\\
&&=\mathbb{E}[\sum_{i_{1}=1,\cdots,i_{N}}\operatorname {Vec}(\mathbf{G}_{n}(i_{n})\otimes\mathbf{G}_{n}(i_{n}))^{\text{T}}
\operatorname{Vec}(\prod_{l\neq n}^{N}(\mathbf{G}_{l}(i_{l})\otimes\mathbf{G}_{l}(i_{l}))]\nonumber\\
&&=\sum_{i_{1}=1,\cdots,i_{N}}\prod_{n=1}^{N}(\mathbf{E}_{R_{n}}\otimes \mathbf{K}_{R_{n}R_{n-1}}\otimes \mathbf{E}_{R_{n-1}}) (\mathbb{E}[(\mathbf{g}_{n}(i_{n})(\mathbf{g}_{n}(i_{n}))^{\operatorname{T}}])
\end{eqnarray}
\end{proof}
where $\mathbf{g}_{n}(i_{n})=\operatorname{Vec}(\mathbf{G}_{n}(i_{n}))$, $\mathbf{E}_{n}\in \mathbb{R}^{n\times n}$ is an unit matrix, $\mathbf{K}_{mn}$ is the permutation matrix.
\end{theorem}

%\subsection{Related Works}
\section{Bayesian Tensor Ring decomposition}
\label{sec:3}

\subsection{Model Description}
In this section, we present the Bayesian low TR rank decomposition based on Student-T process.
Given an incomplete tensor $\mathcal{T}_{\mathbb{O}}\in\mathbb{R}^{I_{1}\times \cdots\times I_{N}}$, its entry is denoted by  $\{\mathcal{T}_{i_{1}i_{2}\cdots i_{N}}|(i_{1},i_{2},\cdots,i_{N})\in\mathbb{O}\}$, where $\mathbb{O}$ is the set of indices of available data in $\mathcal{T}$. Our goal is to find a Bayesian low-TR-rank approximation for the observed tensor $\mathcal{T}$ under probabilistic framework, which is formulated as
\begin{eqnarray}\label{observation}
 && p(\mathcal{T}_{\mathbb{O}}|\{\mathcal{G}_{n}\}_{n=1}^{N},\tau)
  =\prod_{i_{1}}^{I_{1}}\cdots\prod_{i_{N}}^{I_{N}}\\\nonumber
&&\mathcal{N}(\mathcal{T}_{i_{1},i_{2},\cdots,i_{N}}|f(\mathbf{G}_{1}(i_{1}),\cdots,\mathbf{G}_{N}(i_{N})), \tau^{-1})^{\mathcal{O}_{i_{1}\cdots i_{N}}}
\end{eqnarray}
where $\mathcal{N}(x| \mu,\sigma^{2})$ denotes a Gaussian density with mean $\mu$ and variance $\sigma^{2}$, $\tau$ denotes the noise precision, $\mathbf{G}_{n}(i_{n})\in \mathbb{R}^{R_{n-1}\times R_{n}}$ is the $i_{n}$-th slice of $\mathcal{G}_{n}$, and $\mathcal{O}$ is the indicator tensor in which 1 represents observed entry and 0 represents missing entry.
The likelihood model in (\ref{observation}) means the observed tensor is generated by two parts where one is the core factors in TR format and the other is noise.

Firstly, we assume entries of noise are independent and identically distributed random variables obeying Gaussian distribution with zero mean and noise precision $\tau$. To learn $\tau$, we place a hyperprior over the noise precision $\tau$, as follows,
\begin{equation}\label{noise prior}
  p(\tau)=\operatorname{Ga}(\tau|a,b)
\end{equation}
where $\operatorname{Ga}(x|a,b)=\frac{b^{a}x^{a-1}e^{-bx}}{\Gamma(a)}$ denotes a Gamma distribution. The expectation of $\tau$ is defined by $\mathbb{E}[{\tau}]=\frac{a}{b}$.
The parameters $a$ and $b$ are set to small values, e.g., $10^{-7}$, which makes the Gamma distribution a non-informative prior.

Secondly, we assume the recovered tensor has a low rank structure. To learn the low rank structure, we assume the core factor following Student-T distribution and propose a two-layer multiplicative interaction model over core factor. In the first layer, the entries in core factor $\mathcal{G}_{n}$ obey Gaussian distribution with zero mean and a precision matrix:
%However, directly minimizing the tensor ring rank is intractable due to its cycle structure. Therefore, we attempt to seek an automatic selection model which can automatically find the tradeoff between the low tensor ring rank and fitting error. Borrowing the idea from automatic relevance determination~\cite{wipf2008new}, which is to automatically shrink irrelevant variables during inference, we propose a two-layer hierarchical Gaussian prior model. In the first layer, we place a sparse prior on entries of core factors, which is given by:
%we attempt to minimize the latent dimension of each core factors, which corresponding to each entries of core tensor in TR format is sparse.  This probabilistic framework is given by:
 \begin{eqnarray}\label{prior factor}
&&p(\mathcal{G}_{n}|\bm{\lambda}^{(n-1)},\bm{\lambda}^{(n)})=\prod_{i_{n}}^{I_{n}}\prod_{r_{n-1}}^{R_{n-1}}\prod_{r_{n}}^{R_{n}}\\\nonumber
&&\mathcal{N}(\mathcal{G}_{n}(r_{n-1},i_{n},r_{n})|0,(\lambda_{r_{n-1}}^{(n-1)}*\lambda_{r_{n}}^{(n)})^{-1})
\end{eqnarray}
where hyperparameters $\bm{\lambda}^{(n)}=[\lambda_{1}^{(n)},\cdots,\lambda_{R}^{(n)}]$, $\lambda_{r_{n-1}}^{(n-1)}$ and $\lambda_{r_{n}}^{(n)}$ simultaneously control the components in $\mathcal{G}_{n}$.
Specifically,  we directly employ a sparsity-inducing prior over each slice of core factor, leading to the following probabilistic framework:
\begin{eqnarray}
&&p(\mathcal{G}_{n}|\bm{\lambda}^{(n-1)},\bm{\lambda}^{(n)})\\\nonumber
&&=\prod_{i_{n}=1}^{I_{n}}\mathcal{N}(\operatorname{Vec}(\mathbf{G}_{n}(i_{n}))|\bm{0},
(\bm{\Lambda}^{(n-1)}\otimes\bm{\Lambda}^{(n)})^{-1})
\end{eqnarray}
where $\bm{\Lambda}^{(n)}=\operatorname{diag}(\bm{\lambda}^{(n)})$ denotes the precision matrix. The second layer specifies a Gamma distribution as a hyperprior over $\lambda$, as follows:
\begin{equation}\label{lambda}
  p(\bm{\lambda}^{(n)}|\mathbf{c}_{n},\mathbf{d}_{n})=\prod_{r_{n}=1}^{R_{n}}\operatorname{Ga}(\lambda_{r_{n}}^{(n)}|c_{n}^{r_{n}},d_{n}^{r_{n}})
\end{equation}
where the parameters $\mathbf{c}_{n}$ and $\mathbf{d}_{n}$ are set to small values for making Gamma distribution a non-informative prior. Besides, the expectation of $\mathbf{\lambda}^{(n)}$ is defined as:
$$\mathbb{E}[\bm{\lambda}^{(n)}]=\frac{\mathbf{c}_{n}}{\mathbf{d}_{n}}.$$
For simplification, we set $\mathbb{G}=\{\mathcal{G}_{1},\cdots,\mathcal{G}_{N}\}$ and $\bm{\lambda}=\{\bm{\lambda}^{(1)},\cdots,\bm{\lambda}^{(N)} \}$, and all unknown parameters in Bayesian TR model are collected and denoted by $\mathcal{Z}=\{\mathbb{G},\bm{\lambda},\tau\}$.
By combining the stages of the hierarchical Bayesian model, the joint distribution $p(\mathcal{T}_{\mathbb{O}},\mathcal{Z})$ can be written as:
\begin{eqnarray}\label{fullB}
 && p(\mathcal{T}_{\mathbb{O}},\mathcal{Z}) =p(\mathcal{T}_{\mathbb{O}}|\{\mathcal{G}_{n}\}_{n=1}^{N},\tau)p(\tau|a,b)\\\nonumber
&& \times \prod_{n=1}^{N}p(\mathcal{G}_{n}|\bm{\lambda}^{(n-1)},\bm{\lambda}^{(n)})  p(\bm{\lambda}^{(n-1)}|\mathbf{c}_{n-1},\mathbf{d}_{n-1}) p(\bm{\lambda}^{(n)}|\mathbf{c}_{n},\mathbf{d}_{n})
\end{eqnarray}
%To understand clearly, the probabilistic graph model is illustrated in Fig. \ref{Fig:graph}.
%\begin{figure}[ht]
%\vskip 0.2in
%\begin{center}
%\centerline{\includegraphics[width=\columnwidth]{Fig//graph.pdf}}
%  \caption{Probabilistic graphical model of Bayesian TR decomposition.}\label{Fig:graph}
%\end{center}
%\vskip -0.2in
%\end{figure}

Our objective is to compute the conditional density of the latent variables given the observation, as follows:
\begin{equation}
p(\mathcal{Z}|\mathcal{T}_{\mathbb{O}})=\frac{p(\mathcal{Z},\mathcal{T}_{\mathbb{O}})}{\int p(\mathcal{Z},\mathcal{T}_{\mathbb{O}})d_{\mathcal{Z}}}
\end{equation}
Therefore, the missing entries can be inferred by the following equation:
\begin{equation}
p(\mathcal{T}_{\bar{\mathbb{O}}})=\int p(\mathcal{T}_{\bar{\mathbb{O}}}|\mathcal{Z})p(\mathcal{Z})d_{\mathcal{Z}}
\end{equation}
However, the integral over variables $\mathcal{Z}$ is unavailable in closed form, which leads to posterior is intractable to address.
\subsection{Variable Bayesian Inference}
In this section, we apply variable Bayesian inference (VBI) to tackle this problem.
In variational inference, we specify a family $\mathbb{Z}$ of densities over the latent variables. Each $q(\mathcal{Z})\in\mathbb{Z}$
is a candidate approximation to the exact posteriors. Our goal is to find the best candidate, the one closed to $p(\mathcal{Z}|\mathcal{T}_{\mathbb{O}})$ in Kullback-Leibler (KL) divergence, that is:
\begin{equation}\label{opt}
q^{*}(\mathcal{Z})=\argmin_{q(\mathcal{Z})\in\mathbb{Z}}\operatorname{KL}(q(\mathcal{Z}||p(\mathcal{Z}|\mathcal{T}_{\mathbb{O}})))
\end{equation}
 According to KL definition, the problem (\ref{opt}) can be rewritten as:
\begin{equation}\label{opt1}
q^{*}(\mathcal{Z})=\argmin_{q(\mathcal{Z})\in\mathbb{Z}}\mathbb{E}[\ln q(\mathcal{Z})]-\mathbb{E}[\ln p(\mathcal{Z},\mathcal{T}_{\mathbb{O}})]+\ln p(\mathcal{T}_{\mathbb{O}})
\end{equation}
Since $\ln p(\mathcal{T}_{\mathbb{O}})$ is a constant with respect to $q(\mathcal{Z})$, the problem (\ref{opt1}) can be formulated as an optimization model:
\begin{equation}\label{opt2}
\max_{q(\mathcal{G}_{1}),\cdots,q(\mathcal{G}_{N}),q(\bm{\lambda}^{(1)}),\cdots,q(\bm{\lambda}^{(N)}) ,q(\tau)} \mathbb{E}[\ln p(\mathcal{Z},\mathcal{T}_{\mathbb{O}})]-\mathbb{E}[\ln q(\mathcal{Z})]
\end{equation}
where $ q(\mathcal{Z})=q(\tau)\prod_{n=1}^{N}q(\mathcal{G}_{n})\prod_{n=1}^{N}q(\bm{\lambda}^{(n)})
$ based on the mean-filed approximation. It means each parameter in $\mathcal{Z}$ is independent and these parameters could be developed by iteratively optimizing each one while keeping the others fixed.

\subsubsection{Update $q$($\mathcal{G}_{n})$}
By substituting the equation (\ref{fullB}) into the optimization problem (\ref{opt2}), we can obtain the following subproblem with respect to $\mathcal{G}_{n}$:
\begin{equation}\label{opt:Gn}
\max_{q(\mathcal{G}_{n})} \mathbb{E}[\ln p(\mathcal{T}_{\mathbb{O}}|\mathcal{G}_{n},\mathcal{G}^{\neq n},\tau)]+\mathbb{E}[\ln p(\mathcal{G}_{n}|\bm{\lambda}^{(n-1)},\bm{\lambda}^{(n)})  ]-\mathbb{E}[\ln q(\mathcal{G}_{n})]
\end{equation}
where
\begin{equation}
q(\mathcal{G}_{n})=\prod_{i_{n}=1}^{I_{n}}\mathcal{N}(\operatorname{Vec}(\mathbf{G}_{n}(i_{n}))|\tilde{\mathbf{g}}_{n}(i_{n}),\mathbf{V}_{i_{n}}^{n})
\end{equation}
with mean $\tilde{\mathbf{g}}_{n}(i_{n})$ and variance $\mathbf{V}_{i_{n}}^{n}$.

From (\ref{opt:Gn}), we could infer core factor $\mathcal{G}_{n}$ by receiving the messages from the observed data $\mathcal{T}_{\mathbb{O}}$, the rest core factors $\mathcal{G}^{\neq n}$ and the noise precision $\tau$, and incorporating the message from its hyperparameters $\bm{\lambda}^{(n-1)}$ and $\bm{\lambda}^{(n)}$.  By utilizing these messages, the optimization model with respect to each $\mathbf{G}_{n}(i_n), i_n\in\{1,\cdots,I_{n}\}$ could be rewritten as (details of the derivation can be found in sec. 2 of supplemental materials):
\begin{eqnarray}
&& \max_{\tilde{\mathbf{g}}_{n}(i_{n}),\mathbf{V}_{i_n}^{n}}
-\frac{1}{2}\{(\operatorname{Vec}(\mathbf{G}_{n}(i_{n})))^{\operatorname{T}}
(\mathbb{E}[\tau]\mathbb{E}[(\mathcal{G}_{\mathbb{O}_{i_{n}<n>}}^{\neq n})^{\operatorname{T}}\mathcal{G}_{\mathbb{O}_{i_{n}<n>}}^{\neq n}]\nonumber \\
&&+\mathbb{E}[(\bm{\Lambda}^{(n-1)}\otimes\bm{\Lambda}^{(n)})])\operatorname{Vec}(\mathbf{G}_{n}(i_{n}))])\nonumber \\
&&-\mathbb{E}[\tau]\mathcal{T}_{\mathbb{O}_{i_{n}<n>}}^{\operatorname{T}}\mathbb{E}[\mathcal{G}_{\mathbb{O}_{i_{n}<n>}}^{\neq n}]\operatorname{Vec}(\mathbf{G}_{n}(i_{n}))\nonumber\\
&&-\mathbb{E}[\tau]\operatorname{Vec}(\mathbf{G}_{n}(i_{n}))^{\operatorname{T}}\mathbb{E}[(\mathcal{G}_{\mathbb{O}_{i_{n}<n>}}^{\neq n})^{\operatorname{T}}\mathcal{T}_{\mathbb{O}_{i_{n}<n>}}^{\operatorname{T}}]\}\nonumber\\
&&+\frac{1}{2}\{(\operatorname{Vec}(\mathbf{G}_{n}(i_{n})))^{\operatorname{T}}(\mathbf{V}_{i_n}^{n})^{-1} \operatorname{Vec}(\mathbf{G}_{n}(i_{n}))]) \\\nonumber
&&-\tilde{\mathbf{g}}_{n}(i_{n})^{\operatorname{T}}(\mathbf{V}_{i_n}^{n})^{-1}\operatorname{Vec}(\mathbf{G}_{n}(i_{n}))-\operatorname{Vec}(\mathbf{G}_{n}(i_{n}))^{\operatorname{T}}(\mathbf{V}_{i_n}^{n})^{-1}
\tilde{\mathbf{g}}_{n}(i_{n})\}\\\nonumber
\end{eqnarray}

The maximum value reaches
%the approximation posterior distribution $q(\mathcal{G}_{n})$ is updated by:
%\begin{equation}
%q(\mathcal{G}_{n})=\prod_{i_{n}=1}^{I_{n}}\mathcal{N}(\operatorname{Vec}(\mathbf{G}_{n}(i_{n}))|\tilde{\mathbf{g}}_{n}(i_{n}),\mathbf{V}_{i_{n}}^{n})
%\end{equation}
with:
\begin{eqnarray}\label{solution:prior factor}
&&\mathbf{V}_{i_{n}}^{n}=(\mathbb{E}[\tau]\mathbb{E}[(\mathcal{G}_{\mathbb{O}_{i_{n}<n>}}^{\neq n})^{\operatorname{T}}\mathcal{G}_{\mathbb{O}_{i_{n}<n>}}^{\neq n}]+\mathbb{E}[(\bm{\Lambda}^{(n-1)}\otimes\bm{\Lambda}^{(n)})])^{-1}\nonumber\\
&&\tilde{\mathbf{g}}_{n}(i_{n})=\mathbb{E}[\tau]\mathbf{V}_{i_n}^{n}\mathbb{E}[(\mathcal{G}_{\mathbb{O}_{i_{n}<n>}}^{\neq n})^{\operatorname{T}}]\mathcal{T}_{\mathbb{O}_{i_{n}<n>}}.
\end{eqnarray}
%\begin{eqnarray}\label{equation2:corefactor}
%&\ln q(\mathbf{G}_{n}(i_{n}))=
%-\frac{1}{2}(\mathbf{g}_{n}(i_{n}))^{\operatorname{T}}\mathbf{W}(\mathbf{g}_{n}(i_{n}))])\nonumber\\
%&+\mathbb{E}[\tau]\mathcal{T}_{\mathbb{O}_{i_{n}<n>}}\mathbb{E}[\mathcal{G}_{\mathbb{O}_{i_{n}<n>}}^{\neq n}](\mathbf{g}_{n}(i_{n})))+const
%\end{eqnarray}
 %$\mathbf{W}=\mathbb{E}[\tau]\mathbb{E}[(\mathcal{G}_{\mathbb{O}_{i_{n}<n>}}^{\neq n})^{\operatorname{T}}\mathcal{G}_{\mathbb{O}_{i_{n}<n>}}^{\neq n}]+\mathbb{E}[(\bm{\Lambda}^{(n-1)}\otimes\bm{\Lambda}^{(n)})]$, $\mathbf{g}_{n}(i_{n})=\operatorname{Vec}(\mathbf{G}_{n}(i_{n}))\in \mathbb{R}^{R_{n-1}R_{n}\times 1}$,
where
$\mathcal{T}_{\mathbb{O}_{i_{n}<n>}}\in \mathbb{R}^{(I_{n+1}\cdots  I_{N} I_{1}\cdots I_{n-1})_{\mathbb{O}_{i_{n}}}\times i_{n}}$ and  $\mathcal{G}_{\mathbb{O}_{i_{n}<n>}}^{\neq n}\in\mathbb{R}^{ (I_{n+1}\cdots I_{N}I_{1}\cdots I_{n-1})_{\mathbb{O}_{i_{n}}}\times R_{n-1} R_{n}}$, $\mathbb{O}_{i_{n}}$ is the observed entries in $\mathcal{T}_{\mathbb{O}_{i_{n}<n>}}$ and $|O_{i_{n}}|$ is the number of observations $\mathbb{O}_{i_{n}}$.

The main computational complexity of updating core factor comes from the operation of $\mathbb{E}[(\mathcal{G}_{\mathbb{O}_{i_{n}<n>}}^{\neq n})^{\operatorname{T}}\mathcal{G}_{\mathbb{O}_{i_{n}<n>}}^{\neq n}]$. Based on Theorem 2,
we can calculate $\mathbb{E}[(\mathcal{G}_{\mathbb{O}_{i_{n}<n>}}^{\neq n})^{\operatorname{T}}\mathcal{G}_{\mathbb{O}_{i_{n}<n>}}^{\neq n}]$ by the following equation:
\begin{eqnarray}
&&\mathbb{E}[(\mathcal{G}_{\mathbb{O}_{i_{n}<n>}}^{\neq n})^{\operatorname{T}}\mathcal{G}_{\mathbb{O}_{i_{n}<n>}}^{\neq n}]\\\nonumber
&&=\sum_{\mathbb{O}_{i_{n}}}\prod_{l\neq n}^{N}(\mathbf{E}_{R_{l}}\otimes \mathbf{K}_{R_{l}R_{l-1}}\otimes \mathbf{E}_{R_{l-1}}) (\mathbb{E}[(\mathbf{g}_{l}(i_{l})(\mathbf{g}_{l}(i_{l}))^{\operatorname{T}}])
\end{eqnarray}
%where $\mathcal{V}^{\neq n}=\operatorname{TCP}(\mathcal{V}_{n+1},\cdots,\mathcal{V}_{N},\mathcal{V}_{1},\cdots,\mathcal{V}_{n-1})$.
Assuming $I_{n}=I$ and $R_{n}=R$, the computational complexity for the calculation of $\mathbb{E}[(\mathbf{g}_{n}(i_{n})(\mathbf{g}_{n}(i_{n}))^{\operatorname{T}}]
=\tilde{\mathbf{g}}_{n}(i_{n})\tilde{\mathbf{g}}_{n}(i_{n})^{T}+\mathbf{V}_{i_{n}}^{n}$  is $O(R^{6})$. The update of core factor $\mathcal{G}_{n}$ has a complexity of $O((N-1)|{O}_{I}|R^{6})$.
%For updating core factor $\mathcal{G}_{n}$, the complexity needs $O(I^{N}R^{6})$.

\subsubsection{Update $q(\bm{\lambda}^{(n)})$}
Combining equation (\ref{fullB}) with problem (\ref{opt2}), we can obtain the subproblem with respect to $\bm{\lambda}^{(n)}$, as follows (details of the derivation can be found in sec. 3 of supplemental materials):
\begin{eqnarray}\label{equation: lambda}
\max_{q(\bm{\lambda}^{(n)})} &&\frac{1}{2}\{\mathbb{E}[\ln p(\mathcal{G}_{n}|\bm{\lambda}^{(n-1)},\bm{\lambda}^{(n)})+\ln p(\mathcal{G}_{n+1}|\bm{\lambda}^{(n)},\bm{\lambda}^{(n+1)})\nonumber\\
&&+2\ln  p(\bm{\lambda}^{(n)}|\mathbf{c}_{n},\mathbf{d}_{n})]\}-\mathbb{E}[ \ln q(\bm{\lambda}^{(n)})]
\end{eqnarray}
where
 \begin{equation}
 q(\bm{\lambda}^{(n)})=\prod_{r_{n}=1}^{R_{n}}\operatorname{Ga}(\lambda_{r_{n}}^{(n)}|\tilde{c}_{n}^{r_{n}},\tilde{d}_{n}^{r_{n}})
 \end{equation}
 with parameters $\tilde{c}_{n}^{r_{n}}$ and $\tilde{d}_{n}^{r_{n}}$.

As shown in (\ref{equation: lambda}), the inference of $\bm{\lambda}^{(n)}$ can be obtained by receiving the messages from its corresponding core factors, which are $\mathcal{G}_{n-1}$ and $\mathcal{G}_{n}$, and a pair of its partners, including $\bm{\lambda}^{(n-1)}$ and $\bm{\lambda}^{(n+1)}$, meanwhile combining the information with its hyperparameters, which are $\mathbf{c}_{n}$ and $\mathbf{d}_{n}$. Therefore, for each posteriors of $\lambda_{r_n}^{(n)}$, $r_n\in\{1,\cdots,R_{n}\}$, the optimization model is \begin{eqnarray}
&&\max_{\tilde{c}_{n}^{r_{n}}, \tilde{d}_{n}^{r_{n}} }\quad (c_{n}^{r_{n}}+\frac{1}{2}(I_{n}R_{n-1}+I_{n+1}R_{n+1})-1)\ln \lambda_{r_{n}}^{(n)}\nonumber\\
&&-\{(d_{n}^{r_{n}}+\frac{1}{4}(\mathbb{E}[\bm{\lambda}^{(n-1)}]\mathbb{E}[((\mathbf{G}_{n}(r_{n})\mathbf{G}_{n}(r_{n}))^{\operatorname{T}})]\nonumber\\
&&+\mathbb{E}[\bm{\lambda}^{(n+1)}]\mathbb{E}[(\mathbf{G}_{n+1}(r_{n}))^{\operatorname{T}}\mathbf{G}_{n+1}(r_{n})))]\}\lambda_{r_{n}}^{(n)} \nonumber\\
&&-(\tilde{c}_{n}^{r_{n}}-1)\ln \lambda_{r_{n}}^{(n)}+\tilde{d}_{n}^{r_{n}} \lambda_{r_{n}}^{(n)}
\end{eqnarray}
The optimization solutions are obtained by
%
%By using these messages, the distribution of $q(\bm{\lambda}^{(n)})$ can be further developed by:
%
% with
 \begin{eqnarray}\label{solution:rank update}
\tilde{c}_{n}^{r_{n}}&=&c_{n}^{r_{n}}+\frac{1}{2}(I_{n}R_{n-1}+I_{n+1}R_{n+1})\nonumber\\\nonumber
\tilde{d}_{n}^{r_{n}}&=&d_{n}^{r_{n}}+\frac{1}{4}(\mathbb{E}[\bm{\lambda}^{(n-1)}]\mathbb{E}[((\mathbf{G}_{n}(r_{n})\mathbf{G}_{n}(r_{n}))^{\operatorname{T}})]\\
&&+\mathbb{E}[\bm{\lambda}^{(n+1)}]\mathbb{E}[(\mathbf{G}_{n+1}(r_{n}))^{\operatorname{T}}\mathbf{G}_{n+1}(r_{n})))])
\end{eqnarray}
where $\mathbf{G}_{n+1}(r_{n})=\mathcal{G}_{n+1}(r_{n},:,:)\in\mathbb{R}^{I_{n+1}\times R_{n+1}}$, $\mathbf{G}_{n}(r_{n})=\mathcal{G}_{n}(:,:,r_{n})\in\mathbb{R}^{R_{n-1}\times I_{n}}$.

%Each parameter of $\bm{\lambda}^{(n)}$ is an independent Gamma distribution, inferring via (\ref{equation:lambda}):
 %\begin{equation}
% q_{\bm{\lambda}^{(n)}}(\bm{\lambda}^{(n)})=\prod_{r_{n}=1}^{R_{n}}\operatorname{Ga}(\lambda_{r_{n}}^{(n)}|\tilde{c}_{n}^{r_{n}},\tilde{d}_{n}^{r_{n}})
% \end{equation}
% where
%\begin{equation}\label{solution:rank update}
%\tilde{c}_{n}^{r_{n}}=c_{n}^{r_{n}}+\frac{1}{4}\dot{c}_{n}^{r_{n}},\quad\tilde{d}_{n}^{r_{n}}=d_{n}^{r_{n}}+\frac{1}{8}\dot{d}_{n}^{r_{n}}
%\end{equation}
The main computational complexity of updating $\bm{\lambda}^{(n)}$ comes from the calculation of $d_{n}^{r_{n}}$, which can be divided into similar two parts. For one of the parts:
\begin{eqnarray}
&&\mathbb{E}[\bm{\lambda}^{(n-1)}]\mathbb{E}[((\mathbf{G}_{n}(r_{n})\mathbf{G}_{n}(r_{n}))^{\operatorname{T}})]
\\\nonumber
&&=\sum_{i_{n}=1}^{I_{n}}\mathbb{E}[\bm{\lambda}^{(n-1)}](\mathbb{E}[\hat{\mathbf{g}}_{n}(i_{n})]\mathbb{E}[\hat{\mathbf{g}}_{n}(i_{n})]^{\operatorname{T}}+\mathbf{V}_{n}(r_{n}))
\end{eqnarray}
where $\hat{\mathbf{g}}_{n}(i_{n})=\mathcal{G}_{n}(:,i_{n},r_{n})\in \mathbb{R}^{R_{n-1}}$. Therefore, for each $\bm{\lambda}^{(n)}$, the complexity is $O(2 I R^{2})$ under the assumption that all $I_{n}=I$ and $R_{n}=R$.

\subsubsection{Update $q(\tau)$}
 Similarly, the subproblem corresponding to $ \tau$ can be converted into:
\begin{equation}\label{equation:tau}
\max_{q(\tau)} \mathbb{E}[\ln p(\mathcal{T}_{\mathbb{O}}|\{\mathcal{G}_{n}\}_{n=1}^{N},\tau)+\ln p(\tau|a,b)]-\mathbb{E}[\ln q(\tau)].
\end{equation}
where
\begin{equation}
  q(\tau)=\operatorname{Ga}(\tau|\tilde{a},\tilde{b})
\end{equation}
with parameters $\tilde{a}$ and $\tilde{b}$.

Form (\ref{equation:tau}), the inference of $\tau$ can be obtained via receiving messages from observed tensor and core factors, meanwhile, incorporating with the message from the hyperparameters $a$ and $b$. Applying these messages, the (\ref{equation:tau}) could be reformulated as:
\begin{eqnarray}
&& \max_{\tilde{a},\tilde{b}} \quad (a+\frac{O}{2}-1)\ln \tau\nonumber\\
&&-(b+\frac{1}{2}\mathbb{E}[\|\mathcal{O}\odot(\mathcal{T}-f(\mathcal{G}_{1},\mathcal{G}_{2}\cdots,\mathcal{G}_{N}))\|_{\operatorname{F}}^{2}])\tau\nonumber\\
&& -\tilde{a} \ln\tau+\tilde{b}\tau
\end{eqnarray}
%
% Utilizing these messages, the distribution of $q(\tau)$ yields(details of the derivation can be found in sec. 4 of supplemental materials):
%\begin{equation}
%  q(\tau)=\operatorname{Ga}(\tau|\tilde{a},\tilde{b})
%\end{equation}
The maximization value could be obtained
when
\begin{eqnarray}\label{solution:noise update}
&&\tilde{a}=a+\frac{O}{2}\nonumber\\
&&\tilde{b}=b+\frac{1}{2}\mathbb{E}[\|\mathcal{O}\odot(\mathcal{T}-\hat{\mathcal{X}}\|_{\operatorname{F}}^{2}]
\end{eqnarray}
%
%\begin{eqnarray}
%&&\ln q_{\tau}(\tau)=(a+\frac{O}{2}-1)\ln\tau\\\nonumber
%&&-(b+\frac{1}{2}\mathbb{E}[\|\mathcal{O}\odot(\mathcal{T}-f(\mathcal{G})\|_{\operatorname{F}}^{2}])\tau+const
%\end{eqnarray}
where $|O|=\sum_{(i_{1},\cdots,i_{N})\in\mathbb{O}}\mathcal{O}_{i_{1},\cdots,i_{N}}$
is the number of total observations, $\hat{\mathcal{X}}=f(\mathcal{G}_{1},\mathcal{G}_{2},\cdots, \mathcal{G}_{N})$.

It can been seen from (\ref{solution:noise update}), calculating $\tilde{b}$ costs the most time, as follows:
\begin{eqnarray}
&&\mathbb{E}[\|\mathcal{O}\odot(\mathcal{T}-\hat{\mathcal{X}}\|_{\operatorname{F}}^{2}]\\\nonumber
&&=\mathcal{T}_{\mathbb{O}}^{2}-2*\mathcal{T}_{\mathbb{O}}*\hat{\mathcal{X}}_{\mathbb{O}}
+\mathbb{E}[\hat{\mathcal{X}}_{\mathbb{O}}\hat{\mathcal{X}}_{\mathbb{O}}^{\operatorname{T}}]
\end{eqnarray}
with
\begin{equation}
\mathbb{E}[\hat{\mathcal{X}}_{\mathbb{O}}\hat{\mathcal{X}}_{\mathbb{O}}^{\operatorname{T}}]
=\sum_{\mathbb{O}}\prod_{n=1}^{N}(\mathbf{E}_{R_{n}}\otimes \mathbf{K}_{R_{n}R_{n-1}}\otimes \mathbf{E}_{R_{n-1}}) (\mathbb{E}[(\mathbf{g}_{n}(i_{n})(\mathbf{g}_{n}(i_{n}))^{\operatorname{T}}])
\end{equation}
%with
%\begin{eqnarray}
%&&\mathbb{E}[f(\mathcal{G}_{\mathbb{O}})f(\mathcal{G}_{\mathbb{O}})^{\operatorname{T}}]\nonumber\\
%&&=\mathbb{E}[f(\mathcal{G}_{\mathbb{O}})]\mathbb{E}[f(\mathcal{G}_{\mathbb{O}})^{\operatorname{T}}]+\mathbb{E}[f(\mathcal{V})]
%\end{eqnarray}
%where $\mathcal{V}=\operatorname{TCP}(\mathcal{V}_{1},\mathcal{V}_{2},\cdots,\mathcal{V}_{N})$.
we could see the main computational complexity of updating $\tilde{b}$ is $O(|O|NR^{6})$ with all $I_{n}=I$ and $R_{n}=R$.

For clarity, we call this algorithm low TR rank based on VBI framework (TR-VBI) for image completion and summarize it in Algorithm \ref{algo:1}. %The convergence of this algorithm can be monitored by changes in the value of ELBO (details of the derivation can be found in Sec. 5 of online supplemental materials).
\begin{center}
		\begin{algorithm}[tb]
			\caption{TR-VBI algorithm}
			\begin{algorithmic}
				\STATE \textbf{Input}: The observed tensor $ \mathcal T\in\mathcal{R}^{I_{1}\times\cdots\times I_{N}} $,
				index set  $ \mathbb{O}$.
				\STATE \textbf{Initialization}:  $\mathcal{G}_{n}$, $\mathbf{V}_{n}$,  $R_{n}$, $c_{n}^{r_{n}}$, $d_{n}^{r_{n}}$, $\lambda_{r_{n}}^{(n)}=c_{n}^{r_{n}}/d_{n}^{r_{n}}$, $1\leq r_{n}\leq R_{n}$, $1\leq n\leq N$, $a$, $b$, $\tau=a/b$, stopping criterion $ \varepsilon $, the maximum iteration $K$.
				\WHILE{$k\leq K$}
				\STATE  $k=k+1$
				\FOR{$n=1:N$}
				\STATE update the posterior $q(\mathcal{G}_{n})$ via (\ref{solution:prior factor})
				\ENDFOR
                \FOR{$n=1:N$}
				\STATE update the posterior $q(\bm{\lambda}^{(n)})$ via (\ref{solution:rank update})
                \STATE reduce rank $R_{n}$ by eliminating zero-components of $\mathcal{G}_{n}$ and $\mathcal{G}_{n+1}$
				\ENDFOR
				\STATE update the posterior $q(\tau)$ via (\ref{solution:noise update})
				\IF{$\mathbb{E}[\tau]\leq\varepsilon$}
				\STATE break
				\ENDIF
				\ENDWHILE
				\STATE $\textbf{Output:}$  recovered tensor $ \hat{\mathcal{X}} $.
			\end{algorithmic}
			\label{algo:1}
		\end{algorithm}
	\end{center}

\subsection{Discussion}
We have developed an efficient algorithm to automatically determine the TR ranks for tensor completion. From the solution of (\ref{solution:prior factor}), the update of $\mathcal{V}_{n}$ is related with the noise precision $\mathbb{E}[\tau]$, the information from other core factors $\mathbb{E}[(\mathcal{G}_{\mathbb{O}_{i_{n}<n>}}^{\neq n})^{\operatorname{T}}\mathcal{G}_{\mathbb{O}_{i_{n}<n>}}^{\neq n}]$ and the prior $\mathbb{E}[(\bm{\Lambda}^{(n-1)}\otimes\bm{\Lambda}^{(n)})]$. It can be easily inferred the lower the value of noise precision is, the more the information from $\mathbb{E}[(\mathcal{G}_{\mathbb{O}_{i_{n}<n>}}^{\neq n})^{\operatorname{T}}\mathcal{G}_{\mathbb{O}_{i_{n}<n>}}^{\neq n}]$ is. Meanwhile, we could observe the update of noise precision is impacted by fitting error from (\ref{solution:noise update}). Therefore, if the model fits well, there will be more information from other factors than the prior. In addition, from (\ref{solution:rank update}), the update of $\bm{\lambda}^{(n)}$ is associated with its interrelated core factors, which are $\mathcal{G}_{n}$ and $\mathcal{G}_{n+1}$, and its partners $\bm{\lambda}^{(n-1)}$ and $\bm{\lambda}^{(n+1)}$. The values of $\bm{\lambda}^{(n-1)}$ and $\bm{\lambda}^{(n)}$ affect their corresponding core factors. Moreover, the smaller values of $\mathbb{E}[\bm{\lambda}^{(n-1)}]\mathbb{E}[((\mathbf{G}_{n}(r_{n})\mathbf{G}_{n}(r_{n}))^{\operatorname{T}})]
$ and $\mathbb{E}[\bm{\lambda}^{(n+1)}]\mathbb{E}[(\mathbf{G}_{n+1}(r_{n}))^{\operatorname{T}}\mathbf{G}_{n+1}(r_{n})))]$ lead to larger $\bm{\lambda}^{(n)}$. The larger values of $\bm{\lambda}^{(n-1)}$ and $\bm{\lambda}^{(n)}$ will enforce the values in core factor $\mathcal{G}_{n}$ smaller, which will influence the update of $\bm{\lambda}^{(n-1)}$ and $\bm{\lambda}^{(n)}$ in turn. Therefore, this model have a robust capability of automatically adjusting tradeoff between fitting error and TR ranks.
\subsection{Complexity Analysis}
\textbf{Storage Complexity}
For an $N$-order tensor $\mathcal{X}\in \mathbb{R}^{I_{1}\times\cdots\times I_{N}}$, the storage complexity is $\prod_{n=1}^{N}I_{n}$, which increases exponentially with its order. Assuming all $I_{n}=I$ and $R_{n}=R$ in the TR model, we only need to store the core factors and hyperparameters, which are $\mathcal{G}_{1},\cdots,\mathcal{G}_{N}$ and $\bm{\lambda}_{1},\cdots,\bm{\lambda}_{N}$ respectively, leading to $\mathrm{O}(NIR^{2}+NR)$ storage complexity.

\textbf{Computational Complexity}
The computation cost of our proposed algorithm divides into three parts which are the update of core factors $\mathcal{G}_{1},\cdots,\mathcal{G}_{N}$, the update of noise precision $\tau$ operation and the update of hyperparameters $\bm{\lambda}_{1},\cdots,\bm{\lambda}_{N}$. Combining these computation complexities together, the computational complexity of our algorithm is $\mathrm{O}({O}_{I_{n}}(N-1)NR^{6}+2NIR^{2}+ONR^{6})$ for one iteration with $I_{n}=I$ and $R_{n}=R$.
%Besides, the main computational complexity can be reduced to $\max(\mathrm{O}(NOR^{6}), \mathrm{O}(I^{N}R^{6}))$ before the Hadamard product operation with $\mathcal{O}$, where $O$ is number of the total observations.
\section{Experiments}
\label{sec:4}
To evaluate our algorithm TR-VBI, we conduct experiments on synthetic data and real data, and compare it with TR-ALS~\cite{wang2017efficient}, FBCP~\cite{zhao2015bayesian}, HaLRTC~\cite{liu2013tensor} and SiLRTCTT~\cite{bengua2017efficient}. TR-ALS and SiLRTCTT are advanced tensor networks based methods, where the former one utilizes the model based on factorization with the TR ranks known in advance and SiLRTCTT is based on TT rank minimization model, addressed by the block coordinate descent method. FBCP and HaLRTC are based on traditional tensor decompositions, where FBCP uses the CP decomposition in BI framework while HaLRTC explores low Tucker rank structure using ADMM.
%The codes of TR-ALS\footnote{https://github.com/wangwenqi1990/TensorRingCompletion.}, FBCP\footnote{https://github.com/qbzhao/BCPF.}, HaLRTC are available and our code can be found in the online supplemental materials.

 All experiments are tested with respect to different missing ratios (MR), which is:
\begin{equation*}
\text{MR}=\frac{M}{\prod_{n=1}^{N}I_{N}}
\end{equation*}
where $M$ is the number of total missing entries which are chosen randomly in a uniform distribution.

For the experiment on synthetic data, we consider the relative standard error (RSE) as a performance metric. The RSE is defined as
\begin{equation*}
\text{RSE}=\frac{\|\hat{\mathcal{X}}-\mathcal{X}\|_{\text{F}}}{\|\mathcal{X}\|_{\text{F}}}
\end{equation*}
 where $\hat{\mathcal{X}}$ is the recovered tensor and $\mathcal{X}$ is the original one. In addition, peak signal-to-noise ratio (PSNR) are used to evaluate the performance for image recovery experiments too, which is
\begin{equation*}
\text{PSNR}=10\log_{10}(\frac{\text{MAX}_{I}^{2}}{\text{MSE}})
\end{equation*}
where $\text{MAX}_{I}^{2}$ is the possible maximum pixel value of the image, and $\text{MSE}$ is mean squared error between the original image and reconstructed image, which is defined as $\|(\hat{\mathcal{X}}-\mathcal{X})\|/N$.

All tests repeatedly are ran 10 times and accomplished using MatLab 2018a on a desktop computer with 3.30GHz Intel(R) Xeon(R)(TM) CPU and 256GB RAM.
Besides, it is noticed that we assume all initial TR ranks $R_{n}=R, 0\leq n\leq N $ for simplification in the following experiments.
\subsection{Synthetic Data}
In this section, we conduct experiments on 4-order synthetic data $\mathcal{X}\in \mathbb{R}^{I\times I \times I \times I}$ which is generated by the equation (\ref{3}) with a set of core factors $\{\mathcal{G}_{1},\cdots\mathcal{G}_{N}\}$ where $\mathcal{G}_{n}\in \mathbb{R}^{R\times I \times R}$, $R$ is the TR rank, $I$ is the size of dimension and the entries of $\mathcal{G}_{n}$ obey Gaussian distribution. Besides, a tensor with noise can be constructed by adding the noise entries with the clean one, e.g. $\mathcal{Y}=\mathcal{X}+\mathcal{E}$ where $\mathcal{E}$ is a noise tensor with the entry following random Gaussian distribution.

To evaluate the performances of our model, including the rank estimation accuracy and the recovery quality, we consider three groups of experiments on synthetic data in this section. For verifying the rank estimation accuracy, we design two groups of experiments under different conditions. The mean and variance of predictive ranks are utilized to measure the accuracy, which is defined by
\begin{equation*}
\text{AIR}=\frac{1}{10}\sum_{i=1}^{10}\text{mean}(\hat{R}_{i}),
\end{equation*}
\begin{equation*}
\text{Var}=\frac{1}{10}\sum_{i=1}^{10}\text{std}(\hat{R}_{i}),
\end{equation*}
 where AIR and Var represent the mean and the variance respectively, and
%where $\mathbb{IR}$, which is the set of the inferred TR ranks, defined as $\mathbb{IR}=\{\hat{R}_{1},\cdots,\hat{R}_{10}\}$.
each $\hat{R}_{i}, i\in\{1,\cdots,10\}$ are the inferred TR ranks and $i$ is the number of $i$-th tests.
The mean and std functions calculate the mean and variance of $\hat{R}_{i}$ respectively. It is noticed that the rank determination is a success if the $R-0.25\leq$AIR$\leq R+0.25$, where $R$ is the real data rank in this experiment.

\begin{figure}[htbp]
\centering
\subfloat[AIR vs SNR]{
\begin{minipage}[t]{0.5\linewidth}
\centering
\includegraphics[width=1.8in]{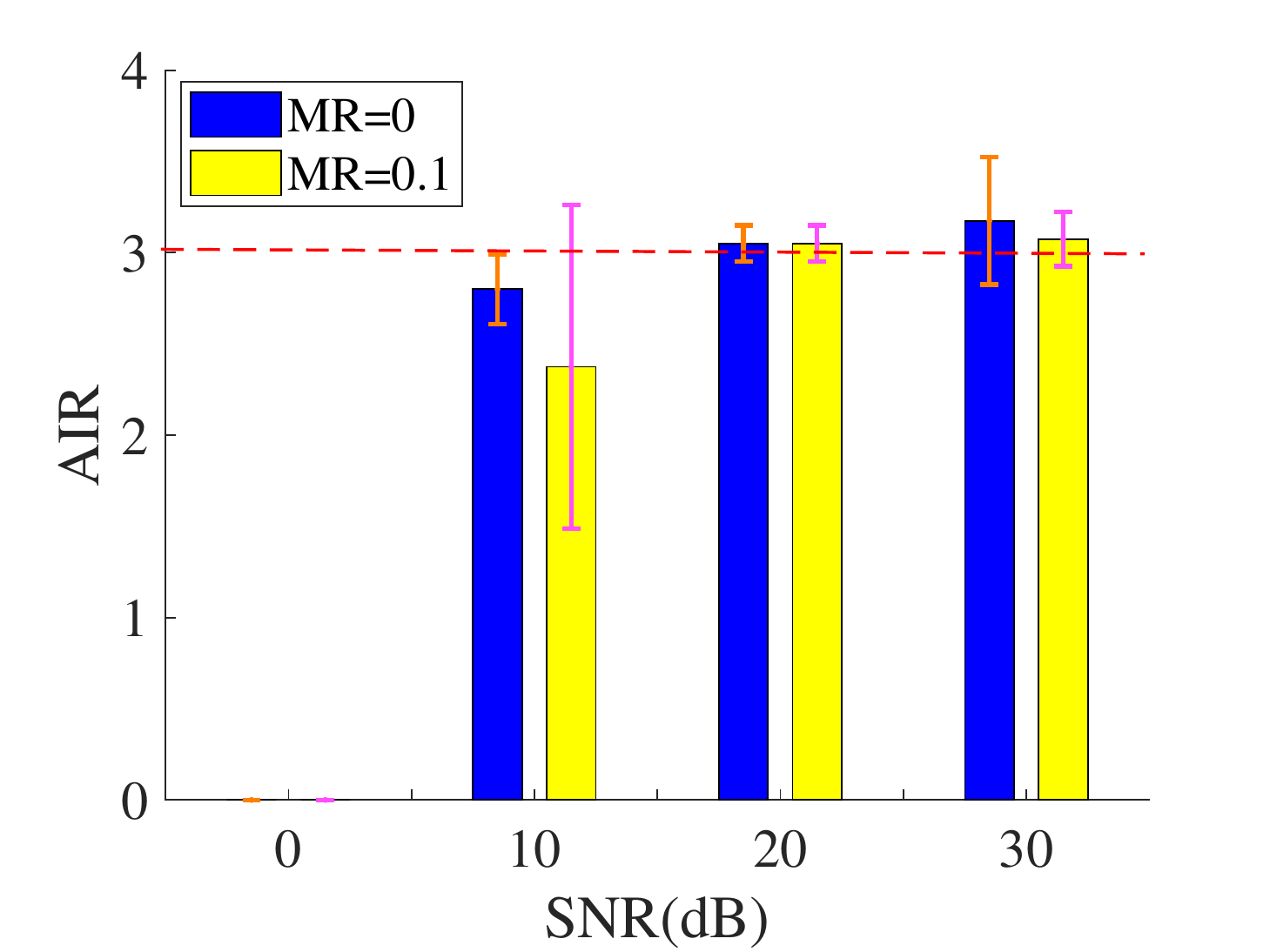}
%\caption{fig1}
\end{minipage}%
}%
\subfloat[AIR vs MR]{
\begin{minipage}[t]{0.5\linewidth}
\centering
\includegraphics[width=1.8in]{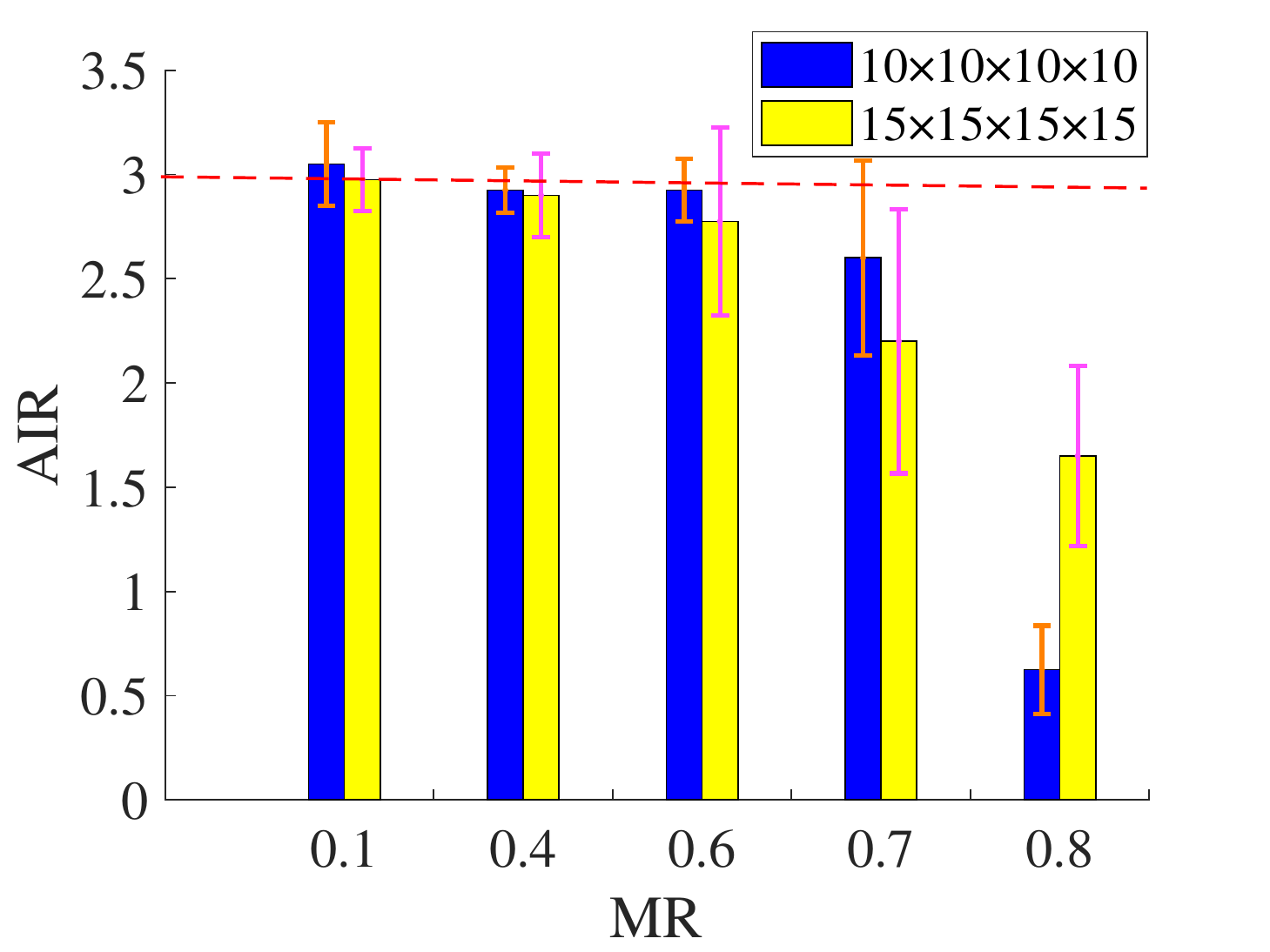}
%\caption{fig2}
\end{minipage}%
}%
\centering
\caption{AIR on different conditions.}
\label{AIR}
\end{figure}

The first group is tested on a 4-order tensor $\mathcal{X} \in \mathbb{R}^{10\times10\times 10\times 10}$ with $R$=3 under different signal noise ratio (SNR) conditions when MR=0.1 and MR=0. The change of AIR and Var along with SNR could be seen in Fig. \ref{AIR}(a). We could observe the inferred rank is reaching the real rank when SNR$\geq$10dB for complete tensor. However, for incomplete tensor with noise, our model can successfully determine the real rank when SNR=20dB.

The second group considered 4-order tensors with $I=10$ and $I=15$, respectively. In this case, SNR=$20$ and $R=3$, and the change of AIR with MR can be illustrated in Fig. \ref{AIR}(b). We could see AIR also tends to the true one with different sizes when MR$\geq$ 0.7.
%From Fig. \ref{AIR}, we could observe the inferred rank is reaching the real rank when SNR$\geq$10dB for complete tensor. If the tensor is incomplete for different size, our model can successful determine the real rank when SNR=20dB and MR. In addition, when SNR=20, AIR also tends to the true one with the size of the observed data changed.
%Accordingly, our proposed model can predict the true rank when SNR$>$10 and the MR is not too large.

  %For data with 10\% missing and no missing, AIR can be obtained when the signal-to-noise ratio is greater than 10 db, and AIR tends to be stable as the SNR increases. In addition, when SNR=20, the size of the observed data is changed, and AIR also tends to the true value. It is therefore possible to predict the size of the exact rank. We can see that when the signal-to-noise ratio is greater than 10db, our proposed model can accurately predict the true rank.
  \begin{figure}[ht]
\begin{center}
\centerline{\includegraphics[width=\columnwidth]{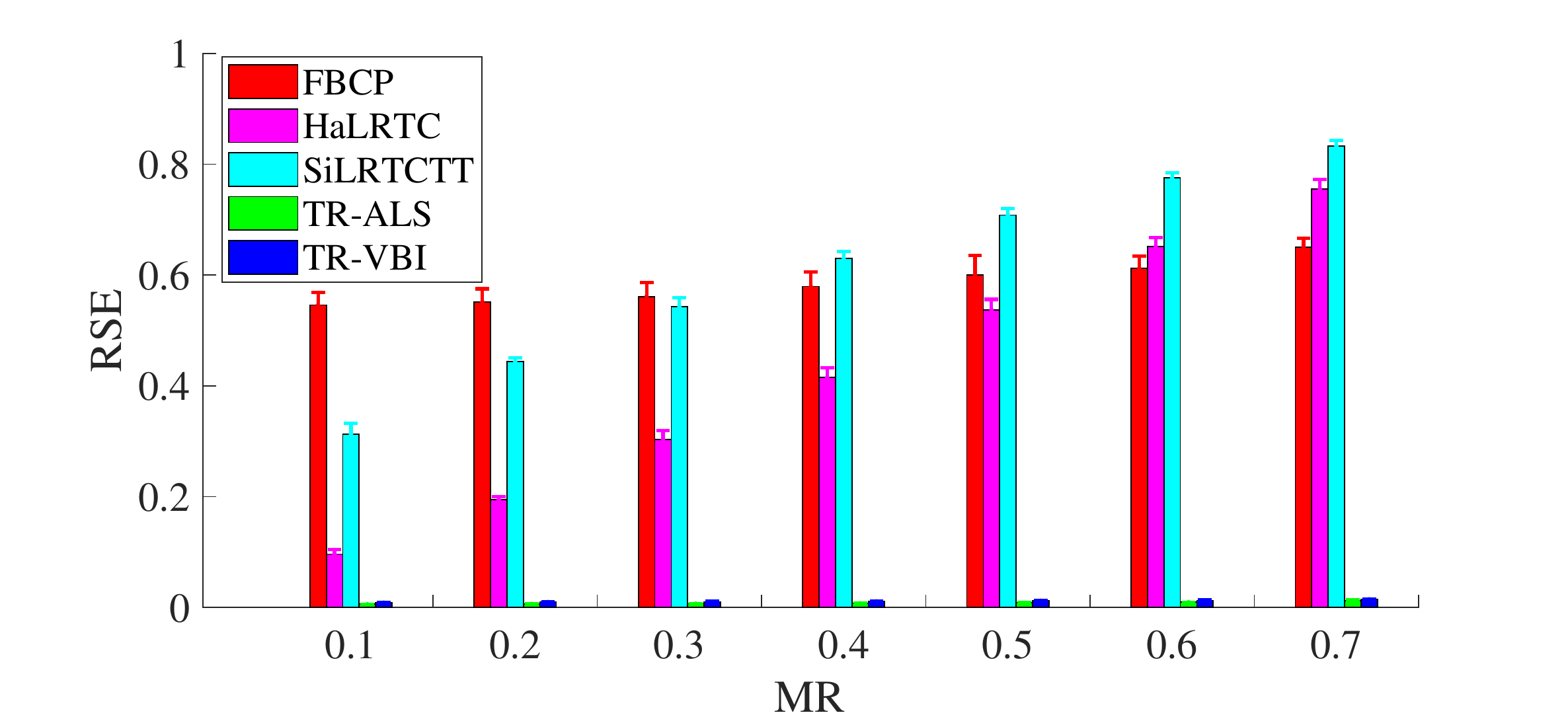}}
  \caption{The comparison of RSE using different methods with MR changed}\label{fig:MRRSE}
\end{center}
\end{figure}
The last one is verified on a 4-D tensor with $I=10$, $R=3$ and SNR=30 using our proposed approach and exiting state-of-the-art methods including TR-ALS, SiLRTCTT, HaLRTC and FBCP. Specifically, TR-ALS with known TR ranks can be seen as a benchmark in this experiment. The result of RSE changing with MR is shown on Fig. \ref{Fig:MRRSE}.
We could see RSE increases with MR growing for all methods. Among these, TR-ALS and TR-VBI approaches outperform others in terms of RSE. Furthermore, the result of TR-VBI is reaching that of TR-ALS with all MRs, which means our proposed algorithm can successfully recover the missing data. On the other hand, the inferred ranks could give a guideline for TR-ALS approach on condition that the  real TR ranks are unknown in advance.

\subsection{Color Images}
\begin{figure}[htbp]
\centering
\subfloat[lena]{
\begin{minipage}[t]{0.23\linewidth}
\centering
\includegraphics[width=.8in]{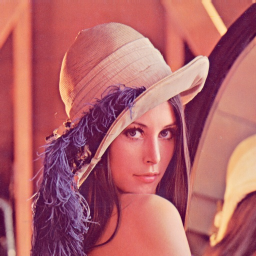}
%\caption{fig1}
\end{minipage}%
}%
\subfloat[Einstein ]{
\begin{minipage}[t]{0.23\linewidth}
\centering
\includegraphics[width=.8in]{fig//natural.png}
%\caption{fig2}
\end{minipage}%
}%
%\quad
\centering
\subfloat[bird]{
\begin{minipage}[t]{0.23\linewidth}
\centering
\includegraphics[width=0.8in]{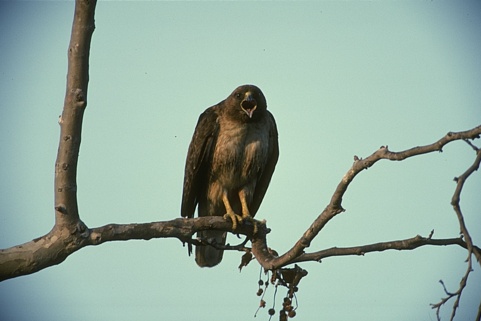}
%\caption{fig1}
\end{minipage}%
}%
\subfloat[dragonfly]{
\begin{minipage}[t]{0.23\linewidth}
\centering
\includegraphics[width=0.8in]{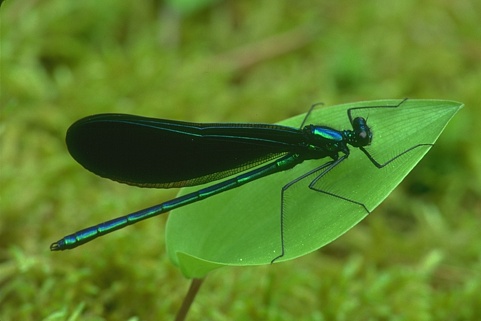}
%\caption{fig2}
\end{minipage}%
}%
\centering
\caption{Testing images with different sizes.}
\label{Testing images}
\end{figure}

In this experiment, we consider the image completion of different sizes of RGB images, including ``lena" with the size $256\times 256\times 3$, ``bird" and `` dragonfly" with the size of $320\times480\times 3$ chosen from Berkeley Segmentation database~\cite{martin2001database}, and ``Einstein" \footnote[1]{https://imgur.com/gallery/5ttQu} with the size of $600\times 600\times 3$. The testing image can be observed in Fig. \ref{Testing images}.

And we compared our proposed one with state-of-the-art methods under noisy (N) and noise-free (NF) conditions with MR ranging from $60\%$ to $95\%$. The SNR we set for ``lena" and ``Esitein" is 20dB and we set SNR=15dB for ``bird" and ``dragonfly". For the traditional decomposition based methods, we consider the parameters $\mathbf{w}=\mathbf{b}/\Arrowvert\mathbf{b}\Arrowvert_{1}$, $\mathbf{b}=[1, 1, 10^{-3}]$ for HaLRTC algorithm followed by~\cite{liu2013tensor}. As suggested in~\cite{zhao2015bayesian}, we set the input rank as 100 for FBCP. For advanced tensor network based ones, literatures~\cite{bengua2017efficient, wang2017efficient,yuan2019high} show casting a low order tensor to a high order without changing the number of entries in the tensor can improve the recovery performance for image completion. Therefore, we reshape a 3-order tensor to a high order for this group, e.g. ``lena" is reshaped as a 9-order tensor $\mathcal{X}\in \mathbb{R}^{4\times 4\times 4\times4\times 4\times 4\times 4\times 4\times 3 }$, ``bird" and ``dragonfly" are reshaped as 9-order tensors with the size $4\times 4\times 4\times5\times 4\times 4\times 5\times 6\times 3$, and ``Einstein" is casted as a 7-order tensor $\mathcal{X}\in \mathbb{R}^{6\times 10\times 10\times 6\times 10\times 10\times 3}$. Following~\cite{bengua2017efficient}, we set the weight parameters $w_{n}=\frac{a_{n}}{\sum_{n=1}^{N-1}a_{n}}$ with $a_{n}=\min(\prod_{l=1}^{n}I_{l},\prod_{l=n+1}^{N}I_{l})$ for SiLRTCTT.  And the entries of initial core factors for our proposed method are randomly chosen from $\mathcal{N}(0,1)$. Besides, we set our inferred ranks as initial TR ranks for TR-ALS.
 \begin{figure*}
 	\centering
 	\includegraphics[width=500pt, keepaspectratio]{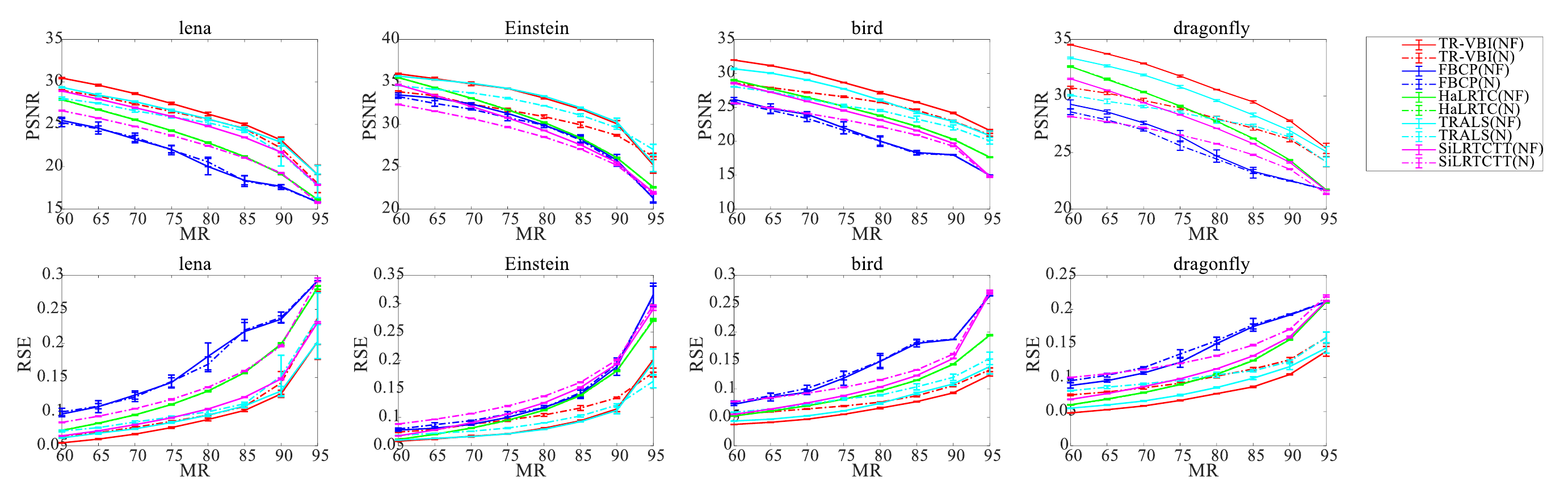}
 	\caption{Performance comparison on image completion using different methods with different MR under different conditions. The first row shows PSNR vs MR; the second row shows RSE vs MR.}
 	\label{fig:PSNR and RSE}	
 \end{figure*}
  \begin{figure*}
 	\centering
 	\includegraphics[width=460pt, keepaspectratio]{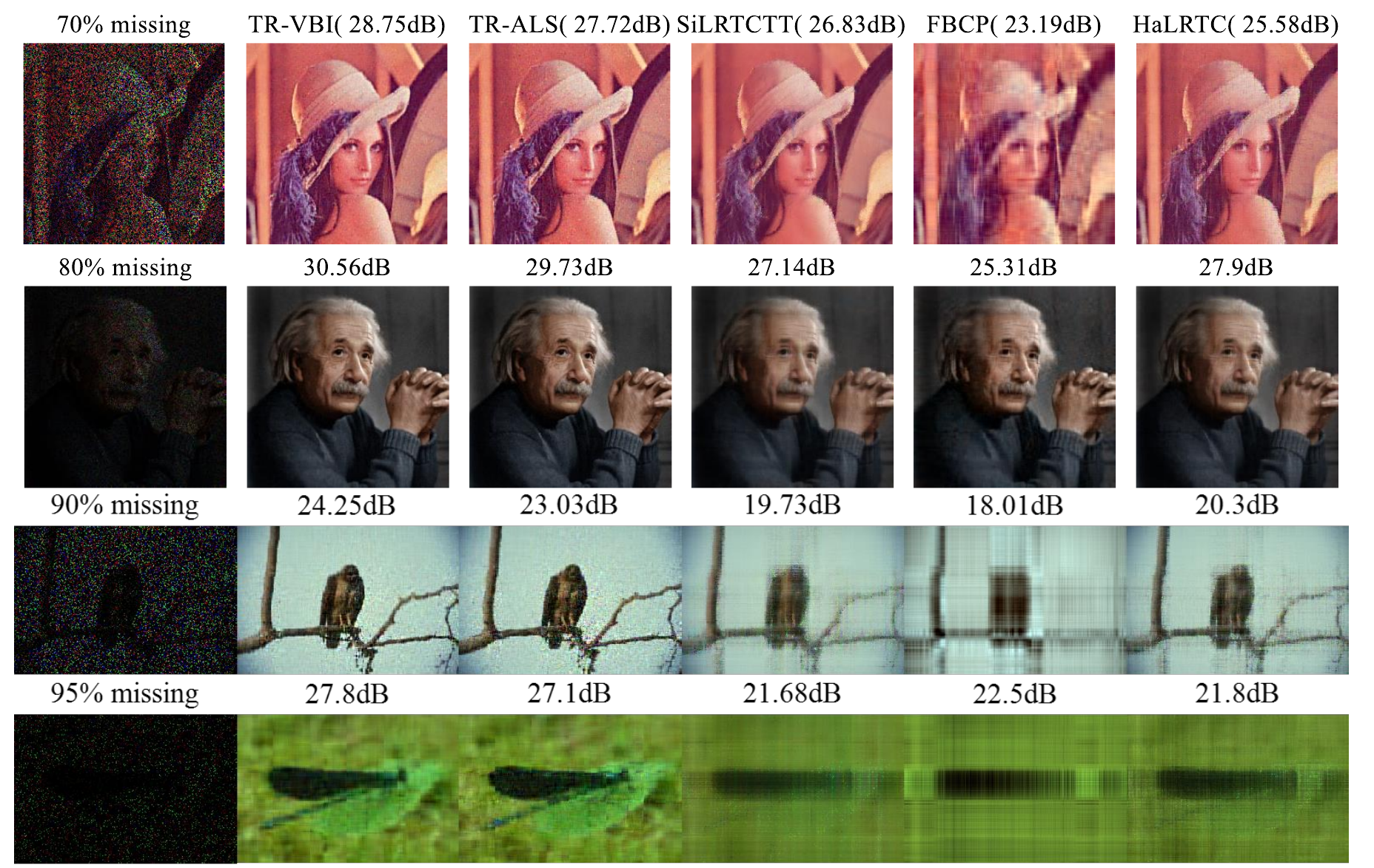}
 	\caption{Examples on image completion using different methods with different MR. }
 	\label{fig:image results1}	
 \end{figure*}

Fig. \ref{fig:PSNR and RSE} shows the performance comparison of different approaches with MR from 60\% ranging to 95\% in terms of PSNR and RSE. Each subfigure illustrates the results by the proposed one and state-of-the-art methods under different conditions where ``N" represents the recovery result with noisy environment and ``NF" indicates the noise-free one. It can be observed the recovery performance of our proposed one outperforms that of others for all tested images under noisy-free condition. Interestingly, the second-best result on image completion is produced by TR-ALS approach. This may imply the TR decomposition can explore more latent information from image. Compared with TR-ALS, the reason that the proposed one achieves better may be our framework has a good ability to balance fitting error and TR ranks. However, the recovery result based on advanced tensor networks  under noisy environment, including TR-VBI, TR-ALS and SiLRTCTT, performs worse than that with noise-free condition.

The recovered results using different algorithms with MR from 70\% to 95\% have been shown in Fig. \ref{fig:image results1}. We could observe recovered images by TR-based methods have a better resolution when MR=90\% and MR=95\%. In addition, TR-VBI and TR-ALS recover more details with MR=70\% and MR=80\%. For example, the light on the hat for ``lena" image and the wrinkles on the forehead for ``Einstein" image are more clear.

\subsection{YaleFace Dataset}
\begin{figure}[ht]
\begin{center}
\centerline{\includegraphics[width=2.5in]{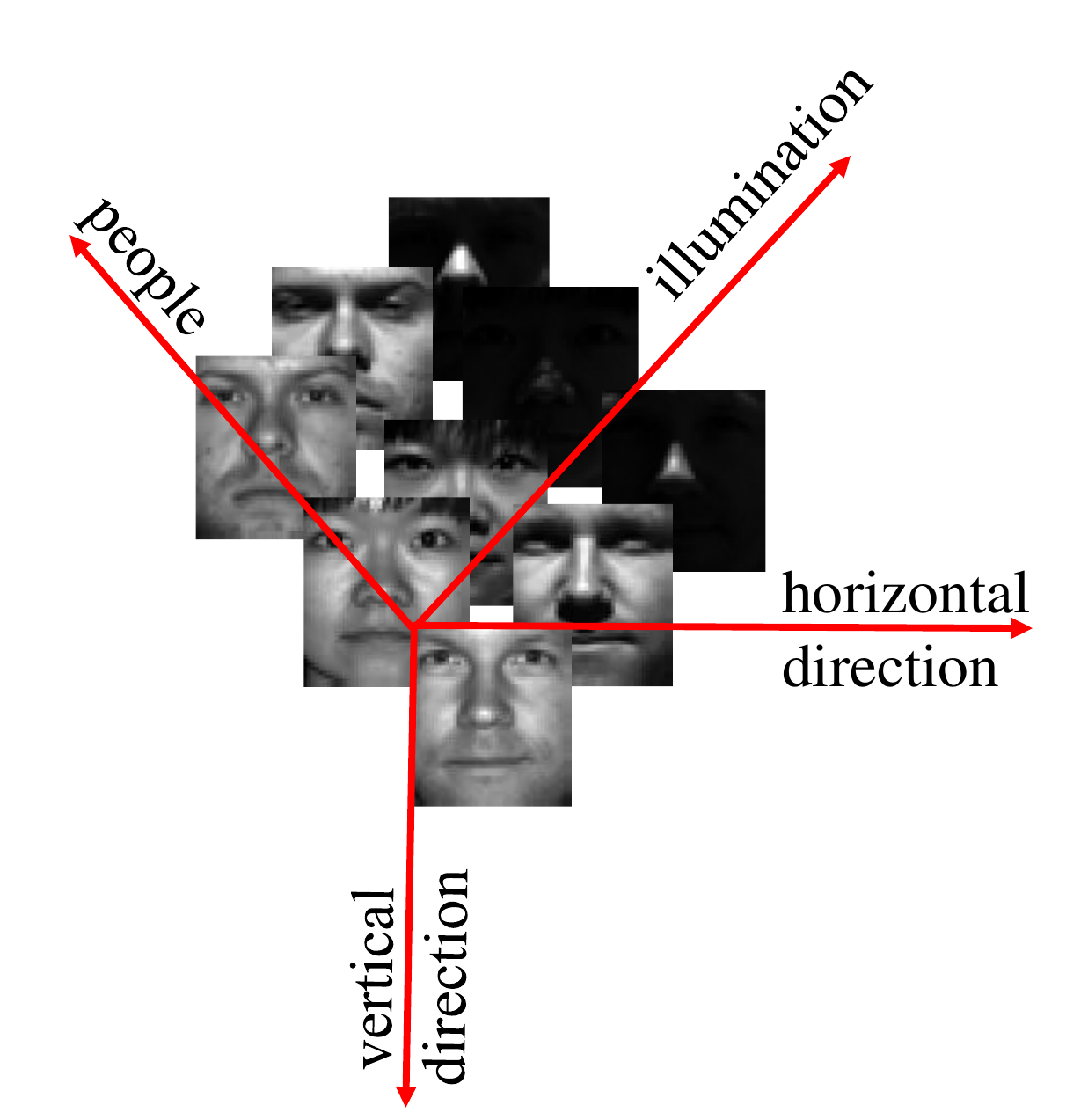}}
  \caption{Extended YaleFace Dataset B with respect to one pose.}\label{Fig:Yale}
\end{center}
\vskip -0.2in
\end{figure}

In this experiment, extended YaleFace Dataset B~\cite{GeBeKr01, KCLee05}, which contains the images of 38 people under 9 poses and 64 illumination conditions where the size of each image is $192\times 168$,  is chosen as a 4D data with respect to one pose ($192\times 168\times 64\times 38$) in this experiment. We downsample the image size to $48\times42$ as a result of computational limitation and reformat the 4D tensor in $\mathbb{R}^{48\times42\times64\times38}$. This can be illustrated in Fig. \ref{Fig:Yale}.
The parameters of the HaLRTCTT are set as $\mathbf{w}=\mathbf{b}/\Arrowvert\mathbf{b}\Arrowvert_{1}$, $\mathbf{b}$=[1, 1, 1, 1]. And the parameters of other methods are also chosen as same as the ones in the color image experiments.

\begin{figure}[ht]
\vskip -0.2in
\begin{center}
\centerline{\includegraphics[width=3in]{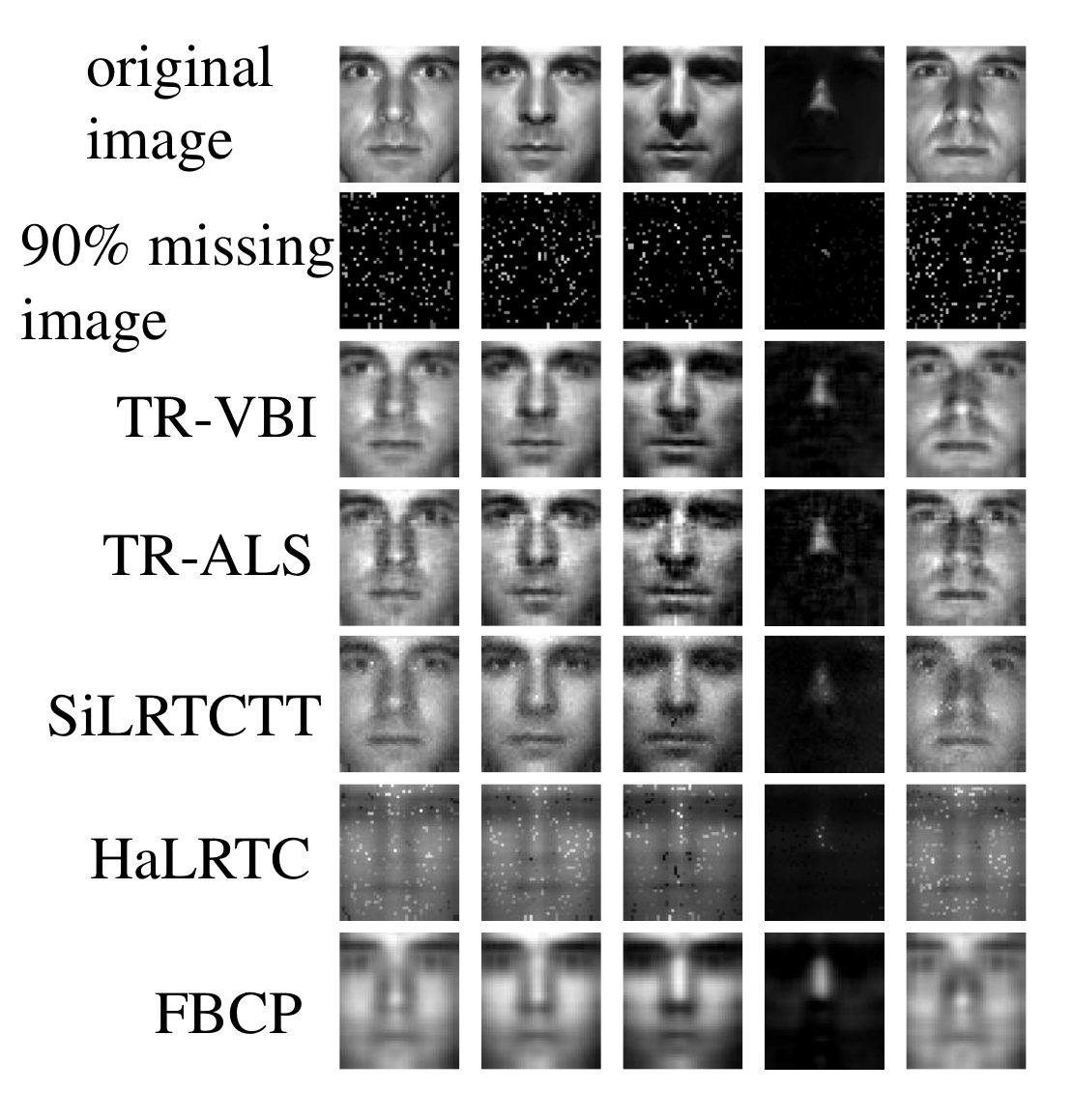}}
  \caption{ The comparison of YaleFace Dataset completion using different approaches when MR=90\%.}\label{Fig:Yale result}
\end{center}
\vskip -0.2in
\end{figure}
Fig. \ref{Fig:Yale result} shows the recovery performance on YaleFace dataset completion by the proposed one and the state-of-the-art methods when MR=90\%. From Fig. \ref{Fig:Yale result}, we could see the methods based on traditional tensor decompositions, including FBCP and HaLRTC, perform worse in terms of recovery quality, which may imply that advanced tensor network based methods explore more information in the high-dimensional data than traditional tensor decompositions based one. Meanwhile, the recovery results of TR-based methods are superior to the one recovered by SiLRTCTT, which may show the advantage of TR decomposition framework on high dimensional data. Furthermore, compared with TR-ALS, TR-VBI could recover the image with a better resolution such as clear eyes and smooth face. Therefore, TR-VBI outperforms all state-of-the-art ones in terms of image recovery performance.

Table \ref{Yalecomplete} illustrates the average results for 10 experiments using different method on YaleFace dataset completion when MR=90\%. We could observe that TR-VBI performs best compared with other approaches in terms of PSNR and RSE. Interestingly, the values of variance in SiLRTCTT and HaLRTC are smaller than that in TR-VBI, TR-ALS and FBCP. This may be caused by the initialization of the latter algorithms which solve problems by iteratively updating core factors.

\begin{table*}[tb]
\vskip 0.15in
\caption{The PSNR/RSE comparision of 90\% missing data via different methods.}
\begin{small}
\begin{center}
\begin{tabular}{cccccc}
 \toprule
  % after \\: \hline or \cline{col1-col2} \cline{col3-col4} ...
    & TR-VBI & TR-ALS & SiLRTCTT & FBCP & HaLRTC \\
    \midrule
  PSNR & \textbf{25.59$\pm$ 0.08} & 24.18$\pm$ 0.08 & 21.90$\pm$ 0.009 & 19.26$\pm$0.33& 16.39$\pm$0.013 \\
    \midrule
  RSE & \textbf{0.1477$\pm$0.0014}& 0.1792$\pm$ 0.002 & 0.2205$\pm$2.2$\times10^{-4}$& 0.30$\pm$0.0113 & 0.4141$\pm$6.29$\times 10^{-4}$ \\
  \bottomrule
\end{tabular}
\end{center}
\end{small}
\vskip -0.1in
\label{Yalecomplete}
\end{table*}

\section{Conclusion}
\label{sec:5}
We have developed a Bayesian low rank TR framework for image completion, which offers a better tradeoff between fitting error and TR ranks. To the best of my knowledge, it is the first time applying TR decomposition in full BI framework on image completion. We utilize mean-field variational inference to approximate the full Bayesian inference and we drive a detailed solution to solve this optimization problem. Several experiments on synthetic data and real-world data demonstrate the superiority of our method over state-of-the-art algorithms.

\ifCLASSOPTIONcaptionsoff
  \newpage
\fi

% trigger a \newpage just before the given reference
% number - used to balance the columns on the last page
% adjust value as needed - may need to be readjusted if
% the document is modified later
%\IEEEtriggeratref{8}
% The "triggered" command can be changed if desired:
%\IEEEtriggercmd{\enlargethispage{-5in}}

% references section

% can use a bibliography generated by BibTeX as a .bbl file
% BibTeX documentation can be easily obtained at:
% http://mirror.ctan.org/biblio/bibtex/contrib/doc/
% The IEEEtran BibTeX style support page is at:
% http://www.michaelshell.org/tex/ieeetran/bibtex/
%\bibliographystyle{IEEEtran}
% argument is your BibTeX string definitions and bibliography database(s)
%\bibliography{IEEEabrv,../bib/paper}
%
% <OR> manually copy in the resultant .bbl file
% set second argument of \begin to the number of references
% (used to reserve space for the reference number labels box)

\bibliographystyle{IEEEtran}
\bibliography{cite}
\label{sec: references}

% that's all folks
\end{document}